\pdfoutput=1

\documentclass[11pt]{article}

\usepackage[preprint]{acl}

\usepackage{times}
\usepackage{latexsym}

\usepackage[T1]{fontenc}

\usepackage[utf8]{inputenc}

\usepackage{microtype}
\usepackage{textcomp}

\usepackage{inconsolata}

\usepackage[utf8]{inputenc} 
\usepackage[T1]{fontenc}    
\usepackage{hyperref}       
\usepackage{url}            
\usepackage{booktabs}       
\usepackage{amsfonts}       
\usepackage{nicefrac}       
\usepackage{microtype}      
\usepackage{xcolor}         
\usepackage{graphicx}
\usepackage{longtable}
\usepackage{array}
\usepackage{pbox}
\usepackage{tabularx}
\usepackage{multirow}
\usepackage{wrapfig}
\usepackage{listings}
\usepackage{pifont}
\usepackage{tcolorbox}
\usepackage{subcaption}
\usepackage{afterpage}
\usepackage{lipsum}

\title{PositionID: LLMs can Control Lengths, Copy and Paste \\ with Explicit Positional Awareness}

\author{
\textbf{Zekun Wang}\textsuperscript{1,2}\thanks{Equal contribution.},\space
\textbf{Feiyu Duan}\textsuperscript{1}\footnotemark[1],\space
\textbf{Yibo Zhang}\textsuperscript{1}\footnotemark[1],\space \\
\textbf{Wangchunshu Zhou}\textsuperscript{3},\space
\textbf{Ke Xu}\textsuperscript{1},\space
\textbf{Wenhao Huang}\textsuperscript{2\thanks{Corresponding author.}},\space
\textbf{Jie Fu}\textsuperscript{4}\footnotemark[2]\\
{\small
\textsuperscript{1}Beihang University;
}
{\small
\textsuperscript{2}01.AI;
} 
{\small
\textsuperscript{3}AIWaves;
} 
{\small
\textsuperscript{4}HKUST;
} \\
\texttt{\small zenmoore@buaa.edu.cn}
}

\begin{document}
\maketitle
\begin{abstract}
Large Language Models (LLMs) demonstrate impressive capabilities across various domains, including role-playing, creative writing, mathematical reasoning, and coding. Despite these advancements, LLMs still encounter challenges with length control, frequently failing to adhere to specific length constraints due to their token-level operations and insufficient training on data with strict length limitations. We identify this issue as stemming from a lack of positional awareness and propose novel approaches—PositionID Prompting and PositionID Fine-Tuning—to address it. These methods enhance the model's ability to continuously monitor and manage text length during generation. Additionally, we introduce PositionID CP Prompting to enable LLMs to perform copy and paste operations accurately. Furthermore, we develop two benchmarks for evaluating length control and copy-paste abilities. Our experiments demonstrate that our methods significantly improve the model's adherence to length constraints and copy-paste accuracy without compromising response quality.\footnote{CP-Bench and LenCtrl-Bench are available in \url{https://huggingface.co/datasets/ZenMoore/CP-Bench} and \url{https://huggingface.co/datasets/ZenMoore/LenCtrl-Bench}.}
\end{abstract}

\section{Introduction}

Large Language Models (LLMs) have demonstrated remarkable capabilities in various domains, such as role-playing~\citep{wang2023rolellm}, creative writing~\citep{wang2024weaver}, mathematical reasoning~\citep{shao2024deepseekmath}, and coding~\citep{roziere2023code}. 
These advanced LLMs often undergo multi-task supervised instruction tuning (SFT), endowing them with strong instruction-following abilities. 
Additionally, an additional alignment training stage after SFT further aligns LLMs with human values and needs. 
A notable outcome of this alignment training is that models tend to produce longer and more detailed responses, thereby enhancing the faithfulness, honesty, and helpfulness of their outputs~\citep{fu2022does, alpaca_eval, dubois2024lcalpacaeval, gpt4}.

However, in many scenarios, longer responses are not necessarily better. 
Users may often prefer outputs that follow specific length constraints or conditions. 
Previous studies indicate that even the most advanced LLMs, such as GPT-4~\citep{gpt4}, struggle to precisely follow length constraints~\citep{instructctg, sun2023evaluating}. 
For example, when users request a model to generate a text strictly within 500 words, the model often produces outputs exceeding this requirement.

We hypothesize that the LLMs' weak adherence to length control instructions can be attributed to two primary issues: 

\begin{itemize}
    \item The model's tokenizer typically operates at the token level rather than the word level, which may mislead the model's perception of the word count. 
    \item The training data may contain a few examples with strict length constraints (\textit{e.g.}, in 3 words), with most data reflecting broad length control requirements, such as ``short'' or ``long''.
\end{itemize}

Intuitively, length control inherently requires an understanding of positional relationships within the text. 
For example, if a model is tasked with generating a summary that is exactly 50 words long, it needs to continuously monitor the number of words it has generated so far and how many words remain to meet the target length. This involves keeping track of its progress (\textit{i.e.}, the number of words being generated) within the text to ensure it stays within the specified limit. 
However, the model's focus on token-level operations and the lack of strict length-constrained training examples impair its ability to effectively and accurately monitor its progress within the text, leading to inaccuracies in following length control instructions. 
We refer to this lack of real-time awareness of the generated text's length by the models as the \textbf{positional awareness issue}.

Our work is not the first to focus on improving the model's positional awareness. 
For example, Abacus Embeddings~\citep{abacus_embedding} enhances the model's mathematical computation capabilities by adding positional embeddings to each digit of a number. 
Similarly, Contextual Position Encoding~\citep{contextual_position_encoding} employs a gate mechanism based on the query-key map to determine positional embeddings, making the positional embeddings context-aware and higher-level. This design improves the model's performance in tasks such as counting and selective copying. 
However, both of them have several limitations: (1) modifying the positional encoding is unfriendly to the off-the-shelf LLMs, as it may degrade their generalist capabilities, and it requires model re-training when altering the position encoding approach. 
(2) And they are not suitable for the closed-source LLMs which only provide invoking APIs. 
Furthermore, (3) these methods are only applicable in limited scenarios, such as some toy tasks or arithmetic tasks, and they demonstrate limited effectiveness when applied to more complex real-world instructions.

To address these issues, we propose novel approaches to enhance the model's positional awareness: PositionID Prompting and PositionID Fine-Tuning for length control tasks. 

In these approaches, we use \textbf{PositionID} to denote the position of each unit in the text, where the unit can be defined as a word, a sentence, a paragraph, etc., according to specific requirements. 
For example, at the word-level, each word is assigned a unique PositionID to indicate its position in the sequence, such as ``The quick brown fox jumps over the lazy dog.'' can be converted into ``The[1] quick[2] brown[3] fox[4] jumps[5] over[6] the[7] lazy[8] dog[9].'' when the position ids are assigned. 
Similarly, at the sentence-level, ``The quick brown fox jumps over the lazy dog. A swift auburn fox leaps across a sleeping canine.'' can be converted into ``The quick brown fox jumps over the lazy dog.[1] A swift auburn fox leaps across a sleeping canine.[2]'' when the position IDs are assigned. 

As shown in Figure \ref{fig:workflow}, 
PositionID Prompting is a tuning-free technique that enables LLMs to count the number of units continuously during text generation. 
PositionID Fine-Tuning involves training the model with data in a similar PositionID Prompting format, thereby enhancing the model's positional awareness. 
Through different system prompts, this method allows flexible control to enable or disable the PositionID prompting mode without compromising positional awareness.

Moreover, we validate another interesting feature brought by explicit positional awareness, namely, copy and paste (CP) abilities. 
To enable this, we propose PositionID CP Prompting, which involves a three-stage tool-use mechanism: 
(1) inserting position ids in the previous text upon generating a ``<COPY>'' token, (2) continuing generating the tool call  ``<COPY>[tag=\(t\)] [desc=\(d\)] [start=\(s\)] [end=\(e\)]</COPY>'' to invoke the copy tool, and (3) generating the tool call ``<PASTE>[tag=\(t\)]</PASTE>'' with position ids eliminated to invoke the paste tool (\textit{c.f.}, \S\ref{positionid-cp}).

To verify the length control (LenCtrl) and copy-paste (CP) abilities of LLMs, we constructed two datasets with diverse instructions: LenCtrl-Bench and CP-Bench. 
LenCtrl-Bench includes a training set of 28,135 samples to enhance the positional awareness of open-source models, and a test set of 2,817 samples. 
CP-Bench contains a test set of 182 samples carefully selected from the GPT-4~\citep{gpt4} synthesized samples and the manually crafted samples. 

Experiments on LenCtrl-Bench demonstrate that through PositionID Prompting and PositionID Fine-Tuning, models can more accurately follow length-controlled text generation instructions without compromising response quality. 
Additionally, our work is the first to study the copy and paste tool-use abilities of current LLMs. 
Experiments on CP-Bench confirm the effectiveness of our proposed PositionID CP Prompting. 
It can not only accurately copy text spans, but also demonstrate superior response quality.

\section{Datasets and Benchmarks}
Due to the lack of instruction datasets and benchmarks specifically designed to evaluate models' abilities to follow instructions with strict length constraints and to perform accurate copying and pasting, we construct two novel benchmarks: LenCtrl-Bench (Length-Controlled Text Generation Benchmark) and CP-Bench (Copy-Paste Benchmark). 
Both of these benchmarks are designed with diverse instructions, aiming to comprehensively evaluate the models' abilities to follow instructions while measuring their length control and copy-paste capabilities.

\subsection{LenCtrl-Bench}

The LenCtrl-Bench is constructed from two well-recognized and large-scale instruction-tuning datasets, namely, Alpaca-52K~\citep{alpaca} and OpenHermes 2.0~\citep{OpenHermes}, along with one additional dataset for text summarization, \textit{i.e.}, WikiHow~\citep{koupaee2018wikihow}. Alpaca-52K and OpenHermes 2.0 involve diverse real-world or GPT-synthesized instructions, but their response lengths are often relatively short. In contrast, due to the nature of text summarization, WikiHow includes longer texts. By combining these three datasets, we can construct a dataset with a broader length distribution.

To evaluate models' length control ability with different levels of length constraints, we design three granularities: (1) word-level, (2) sentence-level, and (3) paragraph-level length constraints. 
We determine the granularity of each sample according to the overall distribution balance and the length of the source data response. 

We use NLTK\footnote{\url{https://www.nltk.org/}} to count the number of words or sentences in the original responses of the source datasets. For WikiHow, we use ``\textbackslash n\textbackslash n'' as a separator to count the number of paragraphs. 
Subsequently, we use GPT-4~\citep{gpt4} to design 24 verbalized instruction templates for length constraints, such as ``With a response length of \{length\} word(s).'', ``Frame your response with \{length\} sentence(s).'', or ``Expand your response to \{length\} paragraph(s).''. 
The length constraint instruction is then concatenated to the original instruction for each sample. 
And we use the original ground-truth responses. 

Moreover, we remove non-English data, data containing code or mathematical expressions, data exceeding 1,000 words in length, and duplicated samples. 
The resulting dataset includes 32,476 samples, which are divided into a training split and a test split with a 10:1 ratio. 
Note that the training split serves as the SFT dataset for PositionID Fine-Tuning when the position ids are assigned for the responses (\textit{c.f.}, \S\ref{sec:postionid-fine-tuning}).

We refer the reader to Appendix \ref{appendix:lenctrl-bench-details} for more details about LenCtrl-Bench, and to Appendix \ref{appx:lenctrl-examples} for examples in LenCtrl-Bench.

\subsection{CP-Bench}

The construction of CP-Bench adopts two approaches: (1) constructing from certain off-the-shelf datasets with frequent textual span repeats, and (2) synthesizing by GPT-4~\citep{gpt4}. 
We brainstorm eight patterns for copy and paste, such as option selection, law article statement, terminology reiteration, note-taking, URL repeating, quotation, policy statement, and general text span repeating. 
The construction approaches, the source datasets, and the descriptions for these CP patterns can be found in Table \ref{tab:cp-patterns}, and their demonstrations are presented in Appendix \ref{appx:cpbench-examples}. 

For Approach (1), we collect the source data from open-source datasets (as detailed in Table \ref{tab:cp-patterns}). 
For each sample in these datasets, we identify repeated sentence blocks (\textit{i.e.}, a sequence of consecutive sentences) and repeated spans under certain regex patterns (\textit{e.g.}, quoted texts, URLs, option contents) as the parts to be copied and pasted. 
We remove samples with too short repeated parts unless they match the aforementioned regex patterns. 

For Approach (2), GPT synthesis, given a CP pattern, GPT-4 is first asked to design 20-40 topics. 
GPT-4 then generates 20 samples for each topic with the repeated spans marked by itself (using square brackets for example). 
The marked repeated spans are the parts to be copied and pasted.

Subsequently, we process these textual forms of the parts to be copied and pasted into tool calling forms. Specifically, for each group of identical parts to be copied and pasted, all instances except for the first occurrence are replaced with a paste tool call, and a copy tool call is inserted before the second occurrence. 
For example, for the text ``My phone number is 1234567. If you miss me, please call 1234567 because 1234567 is my phone number,'' the converted form is ``My phone number is 1234567. If you miss me, please call \{copy\}\{paste\} because \{paste\} is my phone number,'' where \{copy\} and \{paste\} are the tool calls detailed as follows.

Regarding the copy and paste tool calls, the format of \{copy\} is actually ``<COPY> [tag=\(t\)] [desc=\(d\)] [start=\(s\)] [end=\(e\)]</COPY>'', and that of \{paste\} is actually ``<PASTE> [tag=\(t\)] </PASTE>'' correspondingly, where \(t\) and \(d\) denote the textual tags and descriptions for this call, and \(s\) and \(e\) denote the start and end positions for the copied and pasted spans. 
The four tokens, ``<COPY>'', ``</COPY>'', ``<PASTE>'', and ``</PASTE>'', denote the start and end of copy and paste tool calls. 
The tags and descriptions are generated by GPT-4, while the start and end positions are determined by the position ids for the start and end of the original textual spans.

Note that we manually select 182 high-quality samples to create a test set only. We refer the reader to Appendix \ref{appendix:cpbench-detail} for more details about CP-Bench.

\begin{figure*}[t]
    \centering
    \includegraphics[width=\linewidth]{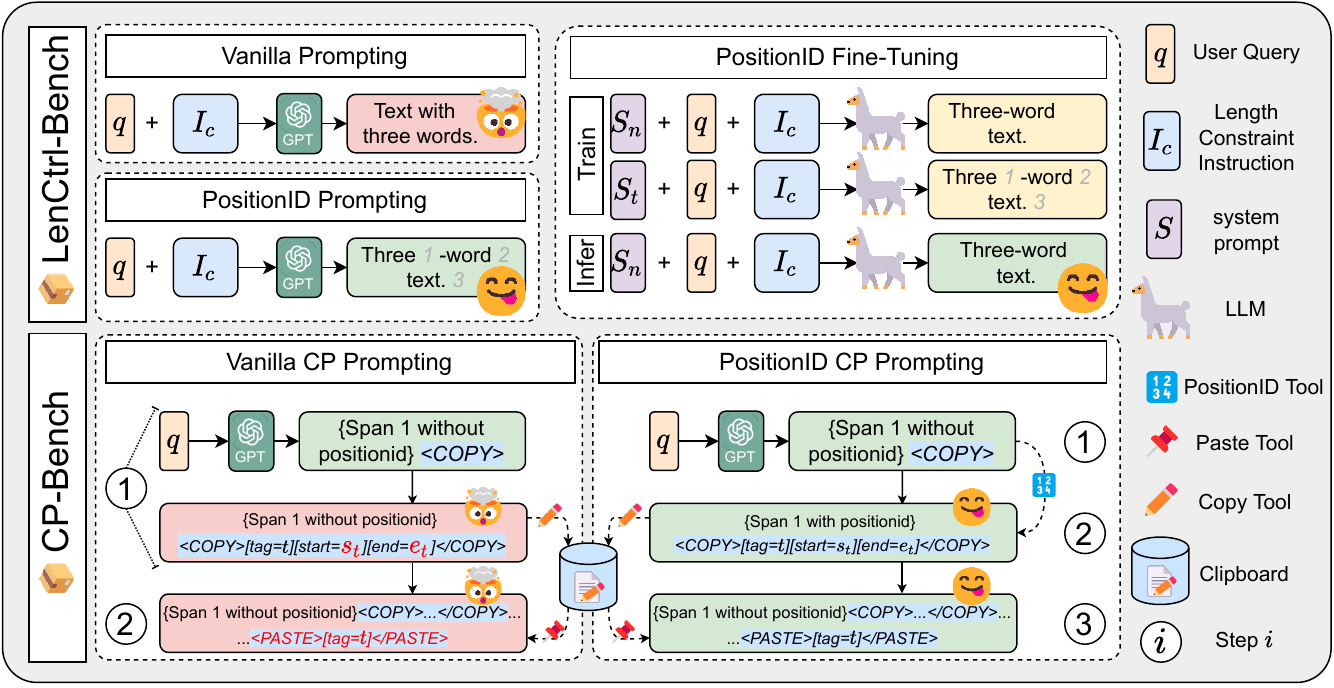}
    \caption{The workflows of our methods. We propose PositionID Prompting and PositionID Fine-Tuning for response length control. \(S_n\) and \(S_t\) denote the system prompts in the normal mode and the PositionID mode, respectively. The model is trained on a mixture of both modes, while inferences are conducted in the normal mode. ``Infer'' denotes ``inference''. Additionally, we introduce PositionID CP Prompting for precise copying and pasting. The font with a blue background indicates tool calls, where \(s\) and \(e\) represent the start and end positions for the copied and pasted spans. The model utilizes external tools to perform copying and pasting operations.}
    \label{fig:workflow}
\end{figure*}

\section{Method}

As shown in Figure \ref{fig:workflow}, we have two pipelines: 
The PositionID Prompting and PositionID Fine-Tuning are designed for length control (\S\ref{positionid-lenctrl}); The PositionID CP Prompting is used for precise copying and pasting (\S\ref{positionid-cp}). 

\subsection{PositionID for Length Control}
\label{positionid-lenctrl}

\paragraph{PositionID Prompting.} 
This approach is primarily suitable for closed-source LLMs such as GPT-4 \citep{gpt4} due to their lack of public availability for training. 
As shown in Figure \ref{fig:workflow}, compared with vanilla prompting, which directly prompts the models with the user query and the length constraint instruction, 
PositionID Prompting elicits the models to generate position IDs for each word, sentence, or paragraph unit using the prompt templates illustrated in Box \ref{appx:prompt-positionid}. 
For example, for a 5-word output, Vanilla Prompting may generate ``The sun is shining brightly.'', while PositionID Prompting generates ``The 1 sun 2 is 3 shining 4 brightly 5.'' with the number after each word defined as the position IDs. 
This is the same case for sentence-level and paragraph-level position IDs. 

This way, the model continuously counts the number of generated words (sentences, or paragraphs) during the next token prediction, enhancing its positional awareness. 
The position signals inherent in the ever-growing context during the inference stage enable the model to continuously and more effectively control the progress of text generation, thereby achieving more strict control over the generated text length.

\paragraph{PositionID Fine-Tuning.\label{sec:postionid-fine-tuning}} 
To enhance positional awareness in open-source models for length control, we propose a novel approach called PositionID Fine-Tuning. 
As shown in Figure \ref{fig:workflow}, we train the language model in a mixture of normal and PositionID modes.

In the normal mode, the system prompt instructs the model to follow standard length control guidelines, such as: ``You are an assistant that strictly follows the length constraint.'' In this mode, the training labels do not include position IDs. 
In contrast, the PositionID mode modifies the system prompt to include positional awareness instructions, such as: ``You are an assistant that generates your response while continuously counting the response length to facilitate length control.'' 
In this mode, the training labels are assigned position IDs.

Training the model in both modes effectively transfers the positional awareness learned in the PositionID mode to the normal mode. 
Consequently, during the inference stage, the system prompt is set to the normal mode to ensure a clean response generation without position IDs.

\begin{table*}[t]
    \centering
    
    \scalebox{1.0}{
    \begin{tabular}{>{\arraybackslash}m{4.5cm}|>{\centering\arraybackslash}m{2.1cm} 
    >{\centering\arraybackslash}m{2.0cm}
    >{\centering\arraybackslash}m{2.2cm}
    >{\centering\arraybackslash}m{2.3cm}
    }
    \toprule
        \textbf{Method} & \textbf{Rouge-L $\uparrow$} & \textbf{MAE (word) $\downarrow$} & \textbf{MAE (sentence) $\downarrow$} & \textbf{MAE (paragraph) $\downarrow$} 
        \\
    \midrule
    \multicolumn{5}{c}{\textit{Length Control Prompting}} \\
    \midrule
        \textbf{Zero-shot} & 22.5 & 8.7 & 2.1 & \underline{2.7} \\ 
        \textbf{Few-shot} & 21.0 & 12.0 & 1.9 & \textbf{2.4} \\ 
        \textbf{CoT} & 22.0 & 17.4 & 1.6 & 3.0 \\ 
        \textbf{Result-informed} & \underline{22.6} & \underline{5.3} & \underline{0.6} & 3.2 \\
        \textbf{Truncation} & \textcolor{gray}{21.6} & \textcolor{gray}{0.0} & \textcolor{gray}{0.0} & \textcolor{gray}{0.0} \\ \midrule
        \textbf{PositionID} & \textbf{23.2} & \textbf{4.8} & \textbf{0.5} & \textbf{2.4} \\ 
    \toprule
    \multicolumn{5}{c}{\textit{Length Control Fine-tuning}} \\
    \midrule
        \textbf{Yi-6B-Chat} & \textbf{20.0} & 58.2 & \underline{5.1} & \textbf{5.9} \\ 
        \textbf{ + CFT} & 18.9 & 60.8 & 5.3 & 6.9 \\ 
        \textbf{ + InstructCTG} & 19.4 & \underline{57.9} & 5.3 & \underline{6.8} \\ \midrule
        \textbf{ + PositionID Fine-Tuning} & \underline{19.8} & \textbf{53.8} & \textbf{4.9} & \underline{6.8} \\
    \bottomrule
    \end{tabular}
    }
    \caption{Experiments on LenCtrl-Bench. We report the Rouge-L score and Mean Absolute Error (MAE) at the word, sentence, and paragraph levels respectively. \textbf{Bold} and \underline{underlined} numbers indicate the best and second-best results. \textcolor{gray}{Gray} scores indicate that the truncation-based baseline does not guarantee the completeness of the response due to response truncation and is therefore omitted in the ranking.}
    \label{tab:eval-prompts}
\end{table*}

\subsection{PositionID for Copy and Paste}
\label{positionid-cp}

\paragraph{PositionID CP Prompting.} 
As shown in Figure \ref{fig:workflow}, PositionID CP Prompting involves a three-step function-calling mechanism that heavily relies on the tool-use capabilities of LLMs~\citep{qin2023toolllm, wang2023interactivenlp}. 
These three steps are (1) the pre-generation phase, (2) the copy tool calling phase, and (3) the paste tool calling phase.

In the pre-generation phase, the LLM generates text up to the token ``<COPY>''. 
Upon generating the ``<COPY>'' token, the LLM stops its text generation process and invokes external tools (\textit{e.g.}, NLTK) to assign position IDs to each word. 
This operation modifies the context for the LLM due to the insertion of numerous position IDs.

During the copy tool calling phase, the visibility of position IDs allows the LLM to generate copy tool calls with precise position parameters, namely, \(s\) and \(e\) in ``<COPY>[tag=\(t\)] [desc=\(d\)] [start=\(s\)] [end=\(e\)]</COPY>''.\footnote{Note that multiple copy tool calls may exist within the context, so the start and end parameters are more precisely referred to as \(s_t\) and \(e_t\).} 
Upon generating the token ``</COPY>'', the external copy tool copies the content between position \(s\) and position \(e\) to an external clipboard, which stores the copied text spans associated with their tags. 
The description parameter in the copy tool calls serves as an auxiliary signal for the LLM to understand the content represented by the copy tool calls, facilitating subsequent pasting.

Finally, in the paste tool calling phase, the model generates paste tool calls, such as ``<PASTE>[tag=\(t\)]</PASTE>'', based on the description parameters of the copy tool calls to paste the copied content tagged with \(t\). 
When the token ``</PASTE>'' is generated, the paste tool executes and replaces the paste tool call span with the corresponding copied content. 
The text generation process then continues until it either ends or encounters the next ``<PASTE>'' or ``<COPY>'' token, thereby repeating the corresponding process.

\section{Experiments}

\subsection{Experimental Setup}
\label{sec:exp_setup}

\paragraph{Baselines} For the \textbf{length control} task, we compare our PositionID prompting with 
(1) zero-shot prompting, which specifies the length constraint in the instruction without using position ids, 
(2) few-shot prompting, which follows the same instruction as zero-shot prompting, but includes three demonstrations in the context, 
(3) CoT prompting, which leverages the Chain of Thought (CoT~\citep{wei2022chain}) techniques for length control. 
CoT Prompting uses an additional prompt: ``Please first generate your thoughts about how to control the length and then generate the content''. 
(4) result-informed prompting, wherein the LLM is prompted with the initial response obtained from zero-shot prompting augmented by the difference between the generated response length and the required length, thereby enabling the model to refine its outputs. 
(5) truncation, which involves suppressing the end-of-sentence token and stopping the model's generation once the specified length is attained. The truncation-based baseline is theoretically stringent in adhering to the length constraint; however, it may not guarantee high-quality content generation. 
We compare our PositionID fine-tuning with 
(1) CFT, which involves constraint-specific fine-tuning with non-verbalized length control attributes~\citep{instructctg}, 
and (2) InstructCTG, which is fined-tuned solely on the normal mode of PositionID Fine-Tuning~\citep{instructctg}. 

For the \textbf{copy and paste} task, we compare our PositionID CP prompting with the subsequent baselines: (1) zero-shot non-CP prompting: the model directly generates responses following the instruction. (2) few-shot CP prompting: similar to PositionID CP Prompting, without inserting PositionID when using the copy tool.

\begin{table}[t]
    \centering
    \scalebox{1.0}{
    \begin{tabular}{>{\arraybackslash}m{1.8cm}|>{\centering\arraybackslash}m{1.2cm} 
    >{\centering\arraybackslash}m{1.6cm}
    >{\centering\arraybackslash}m{1.3cm}
    }
    \toprule
        \textbf{Method} & \textbf{R-L $\uparrow$} & \textbf{CP S.R. $\uparrow$} & \textbf{PPL $\downarrow$} \\
    \midrule
        \textbf{Zero-shot} & 16.3 & \textbackslash{} & 23.1 \\
        \textbf{Few-shot} & 17.4 & 68.1 & 13.5 \\
        \textbf{PositionID} & \textbf{18.4} & \textbf{80.8} & \textbf{8.4} \\
    \bottomrule
    \end{tabular}
    }
    \caption{Experiments on CP-Bench, with results of objective metrics. "R-L": Rouge-L. "S.R.": Success Rate.}
    \label{table:cp-bench-object}
\end{table}

\begin{table}[t]
    \centering
    \scalebox{0.7}{
    \begin{tabular}{>{\arraybackslash}m{2.2cm}|>{\centering\arraybackslash}m{1.5cm} 
    >{\centering\arraybackslash}m{1.5cm}
    >{\centering\arraybackslash}m{1.5cm} 
    >{\centering\arraybackslash}m{2cm}
    }
    \toprule
        \textbf{Method} & \textbf{Clarity} & \textbf{Accuracy} & \textbf{Consistency} & \textbf{Tool-Use Proficiency} \\
    \midrule
        \textbf{Few-shot} & 11.6 & 41.9 & 22.1 & 2.4 \\
        \textbf{PositionID} & \textbf{33.7} & \textbf{55.8} & \textbf{46.5} & \textbf{3.5} \\
    \bottomrule
    \end{tabular}
    }
    \caption{Experiments on CP-Bench, with results of subjective metrics. For Clarity, Accuracy, and Consistency, we report the win rate; for Tool-use proficiency, we report the average scores assigned by GPT-4o.}
    \label{table:cp-bench-subject}
\end{table}

\begin{table}[t]
    \centering
    \scalebox{0.9}{
    \begin{tabular}{>{\arraybackslash}m{2cm}|>{\centering\arraybackslash}m{0.9cm} 
    >{\centering\arraybackslash}m{1.1cm}
    >{\centering\arraybackslash}m{1.1cm} 
    >{\centering\arraybackslash}m{1.1cm}
    }
    \toprule
        \textbf{PositionID Granularity} & \textbf{R-L} & \textbf{MAE (word)} & \textbf{MAE (sent.)} & \textbf{MAE (para.)} \\ 
    \midrule
        \textbf{Word} & 22.4 & \textbf{4.8} & 3.5 & 7.6 \\ 
        \textbf{Sentence} & 23.1 & 16.6 & \textbf{0.5} & 8.8 \\ 
        \textbf{Paragraph} & 22.3 & 21.0 & 2.3 & \textbf{2.4} \\ 
    \bottomrule
    \end{tabular}
    }
    \caption{Comparison of PositionID prompting with different levels of granularity on the LenCtrl-Bench. "R-L": Rouge-L. "sent.": sentence. "para.": paragraph.}
    \label{table:ablation-on-prompt}
\end{table}

\paragraph{Evaluation metrics}

In terms of the length control task, we employ \textbf{Rouge-L} to assess the quality of the generated responses, and the \textbf{Mean Absolute Error (MAE)} to evaluate the difference between the length of the model's generated response and that required by the instructions. We evaluate on three levels of granularity: word-level, sentence-level, and paragraph-level. 

For the copy-and-paste task, we adopt \textbf{Rouge-L} and \textbf{perplexity (PPL)} to evaluate the accuracy and fluency of the generated responses. Additionally, we employ \textbf{CP Success Rate}, which assesses if the models can successfully use the copy and paste tools. 
We also use GPT-4o to assess the following aspects: (1) \textbf{Clarity}, which refers to the model's ease of understanding, smoothness, and the absence of any spelling or grammatical errors. (2) \textbf{Accuracy}, which refers to the correctness and precision of the information, ensuring that all content provided is factually true and exact. (3) \textbf{Consistency} of key information, meaning that all critical elements—such as quoted text, URLs, specific sentences from the original text, provided options, or technical terms—are identical and copied accurately throughout the response. (4) \textbf{Tool-use Proficiency}, which refers to the overall performance of utilizing copy-and-paste tools, ensuring that the copied content is useful.

See Appendix \ref{appx:evaluation} for further details on metrics.

\paragraph{Experiment Details}

For the prompting tasks, all our experiments are conducted using GPT-4-Turbo ~\citep{gpt4}. We utilize the OpenAI official API, setting the temperature parameter to 0.5. For fine-tuning, we employ Yi-6B-Chat ~\citep{young2024yi}. We refer the reader to Appendix \ref{appendix:implementation_details} for additional details.

\begin{figure*}[t]
\centering

\begin{subfigure}[b]{0.24\textwidth}
    \includegraphics[width=\textwidth]{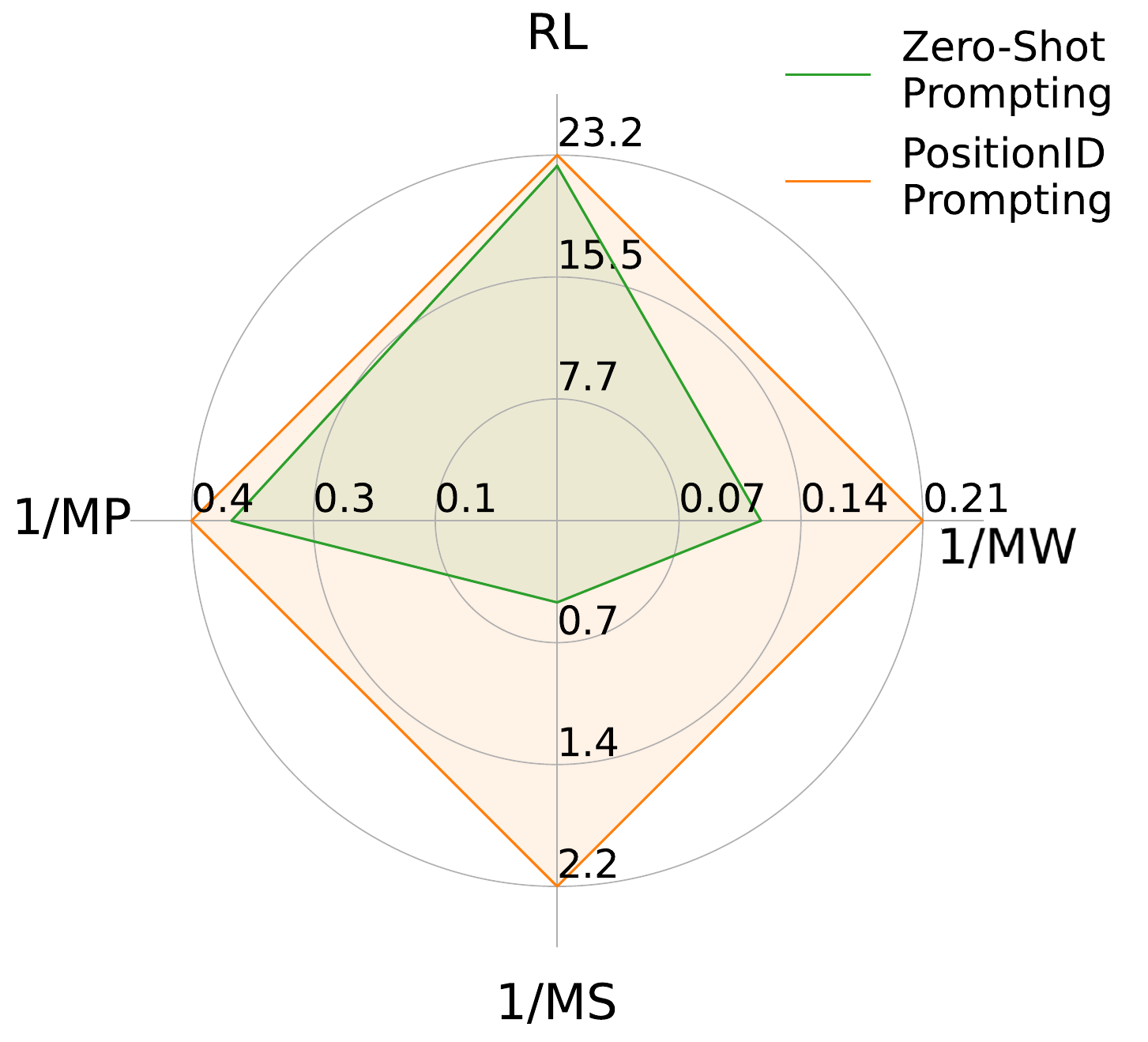}
    \caption{GPT-4}
\end{subfigure}
\begin{subfigure}[b]{0.24\textwidth}
    \includegraphics[width=\textwidth]{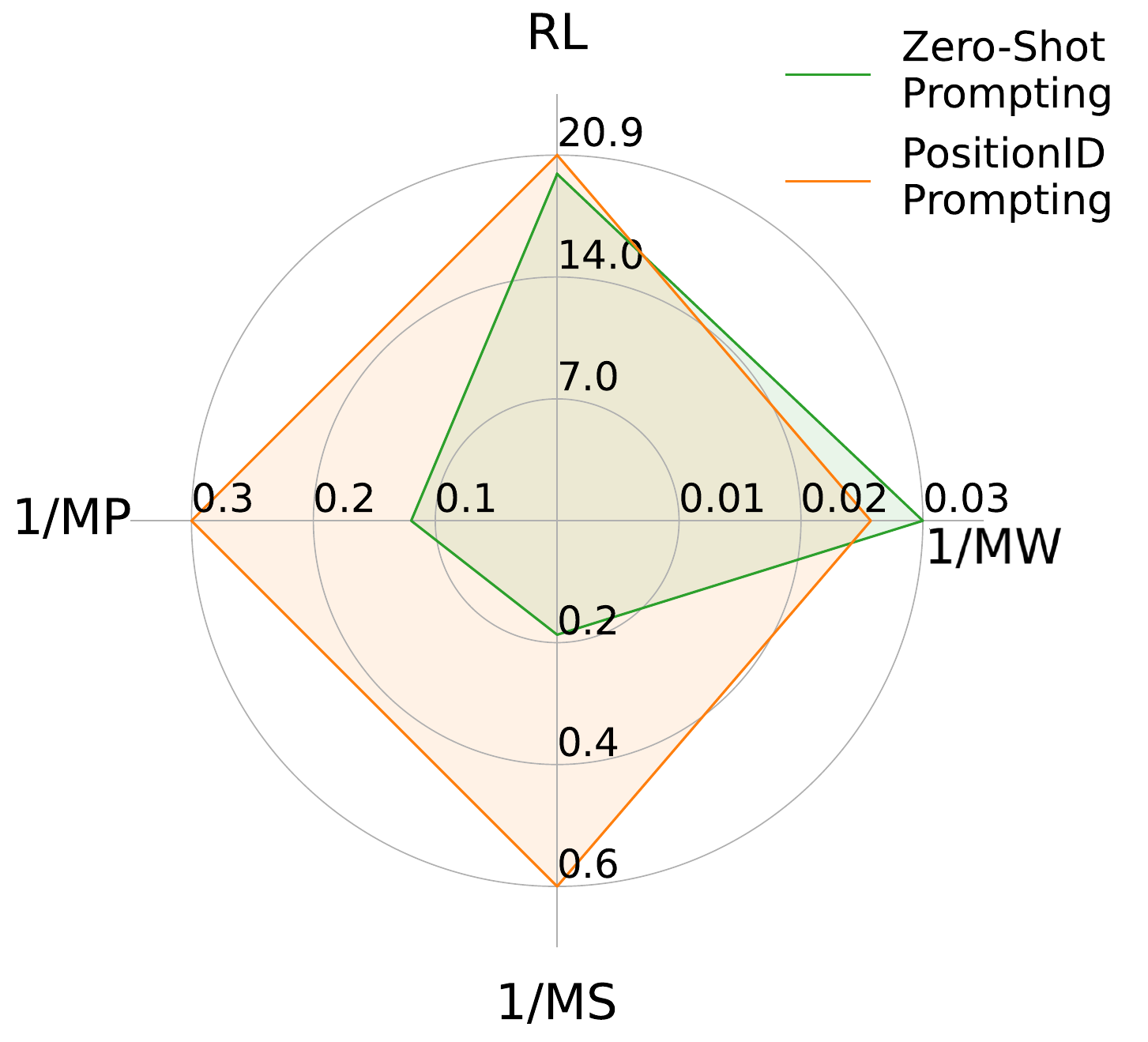}
    \caption{Mistral-7B-Instruct}
\end{subfigure}
\begin{subfigure}[b]{0.24\textwidth}
    \includegraphics[width=\textwidth]{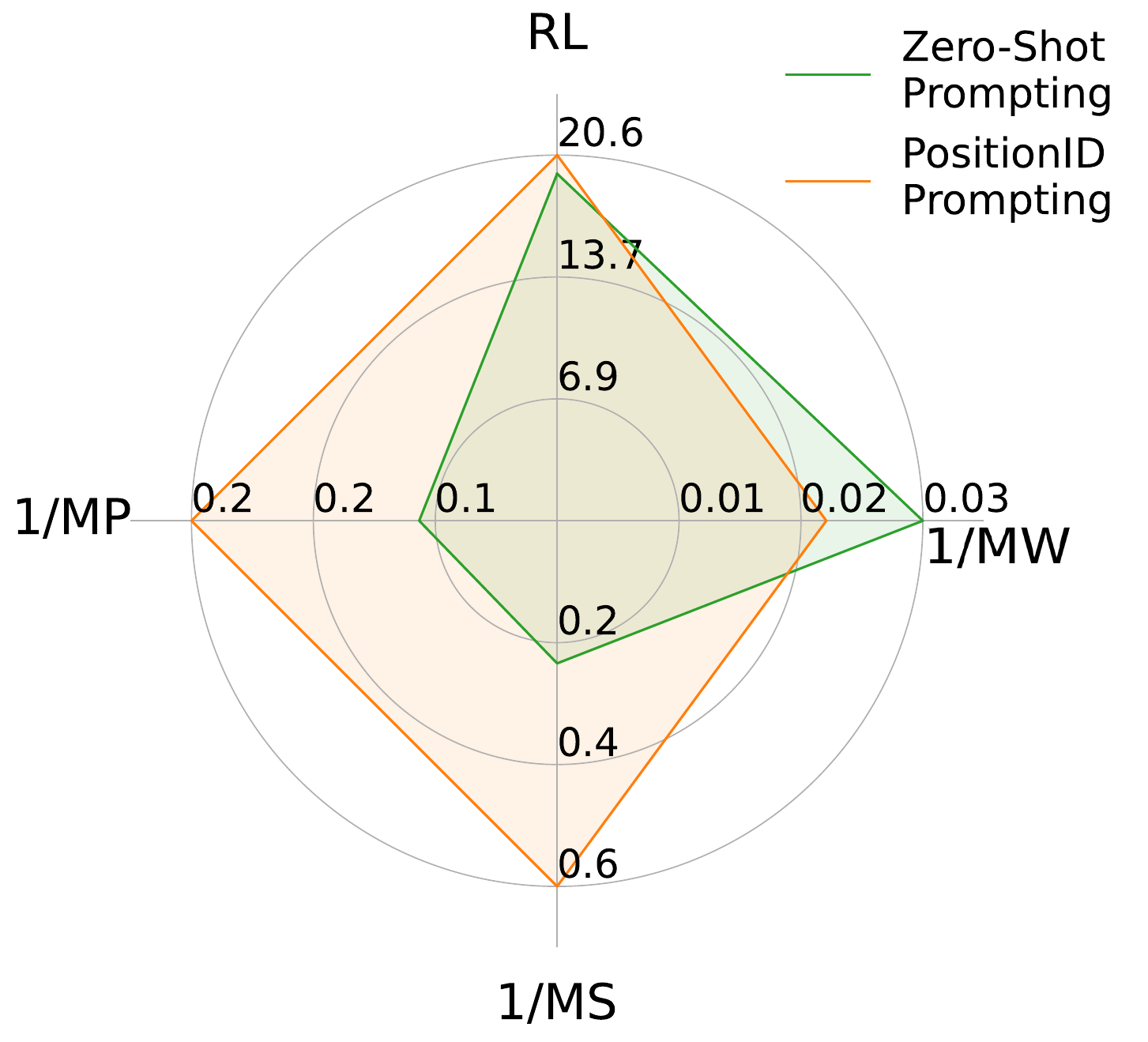}
    \caption{Vicuna-7B}
\end{subfigure}
\begin{subfigure}[b]{0.24\textwidth}
    \includegraphics[width=\textwidth]{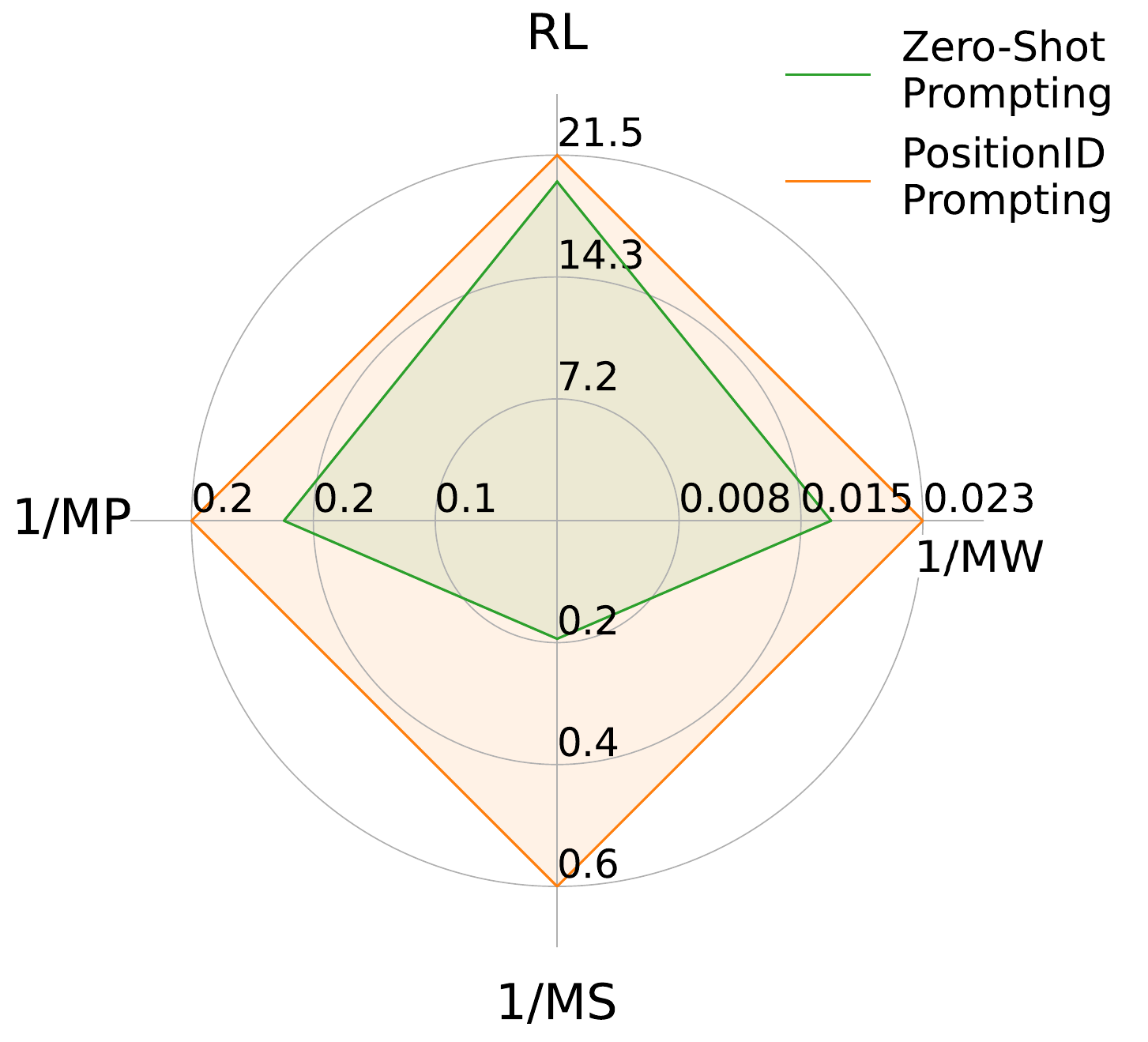}
    \caption{Yi-6B-Chat}
\end{subfigure}

\caption{Comparing PositionID Prompting with Zero-Shot Prompting on LenCtrl-Bench. ``RL'', ``MW'', ``MS'', and ``MP'' denote ``Rouge-L ($\uparrow$)'', ``MAE (word) ($\downarrow$)'', ``MAE (sentence) ($\downarrow$)'', and ``MAE (paragraph) ($\downarrow$)'', respectively.}
\label{fig:Different_Models}
\end{figure*}

\begin{figure*}[h]
\centering

\begin{subfigure}[b]{0.32\textwidth}
    \includegraphics[width=\textwidth]{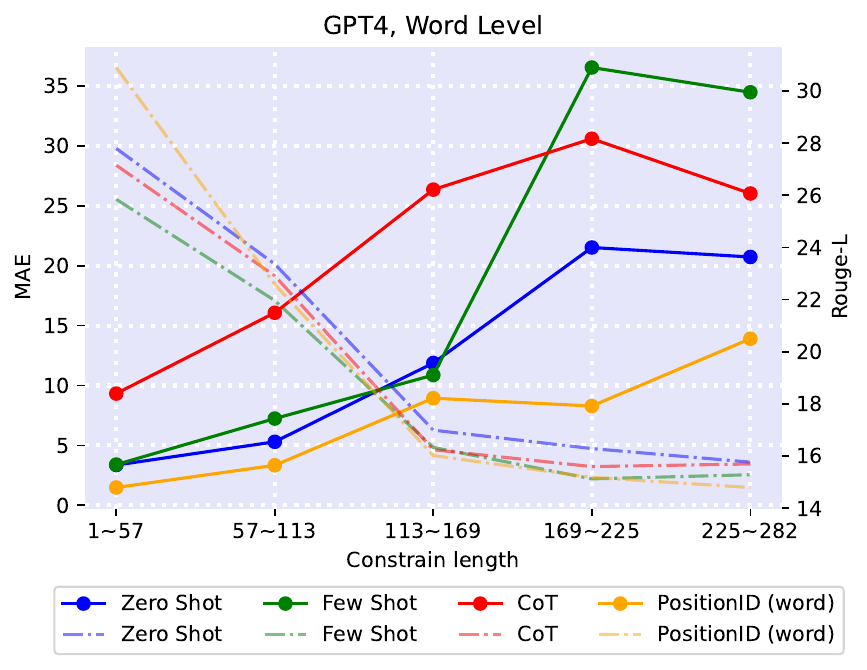}
    \caption{}
\end{subfigure}
\hfill
\begin{subfigure}[b]{0.32\textwidth}
    \includegraphics[width=\textwidth]{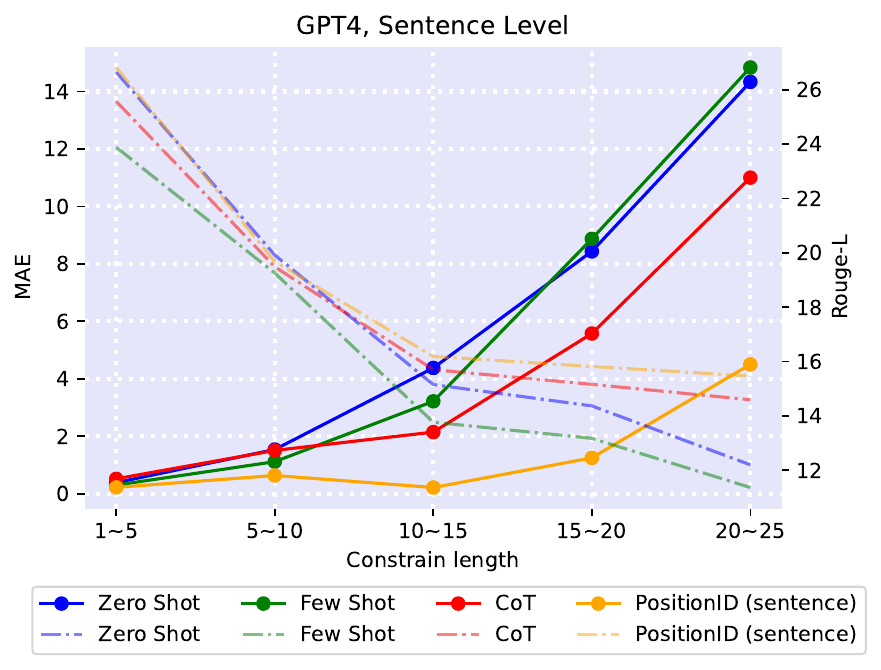}
    \caption{}
\end{subfigure}
\hfill
\begin{subfigure}[b]{0.32\textwidth}
    \includegraphics[width=\textwidth]{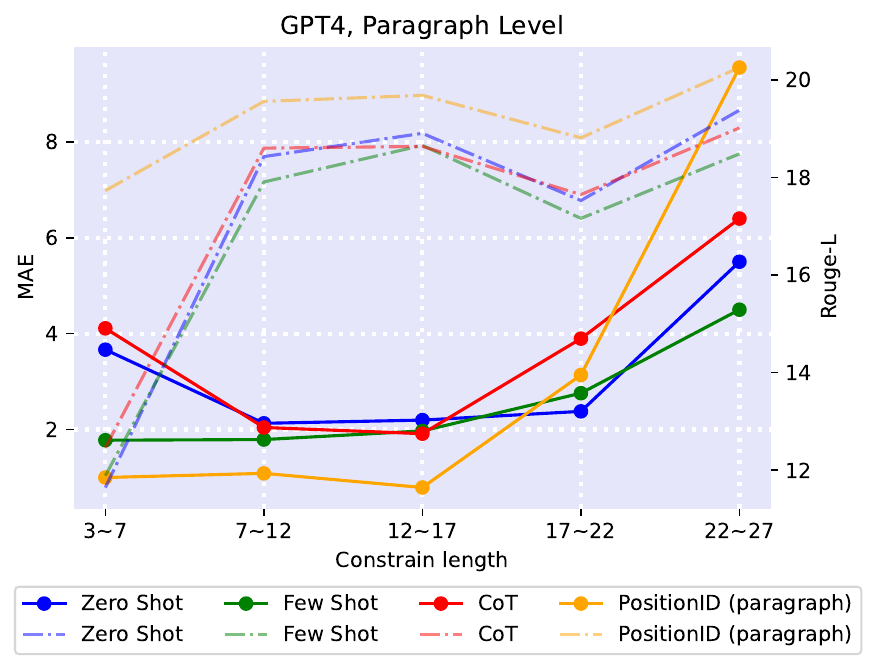}
    \caption{}
\end{subfigure}

\vspace{1em} 

\begin{subfigure}[b]{0.32\textwidth}
    \includegraphics[width=\textwidth]{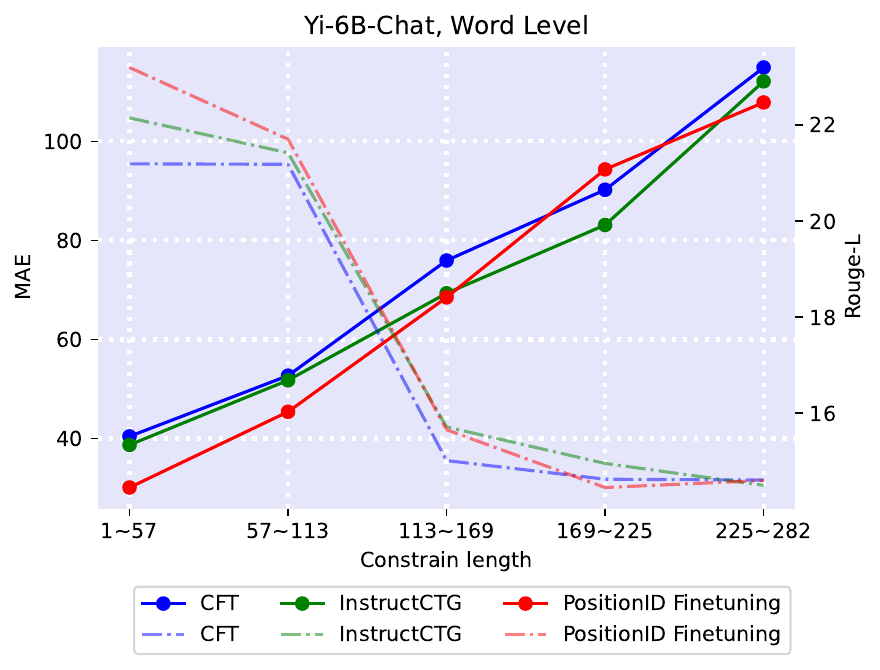}
    \caption{}
\end{subfigure}
\hfill
\begin{subfigure}[b]{0.32\textwidth}
    \includegraphics[width=\textwidth]{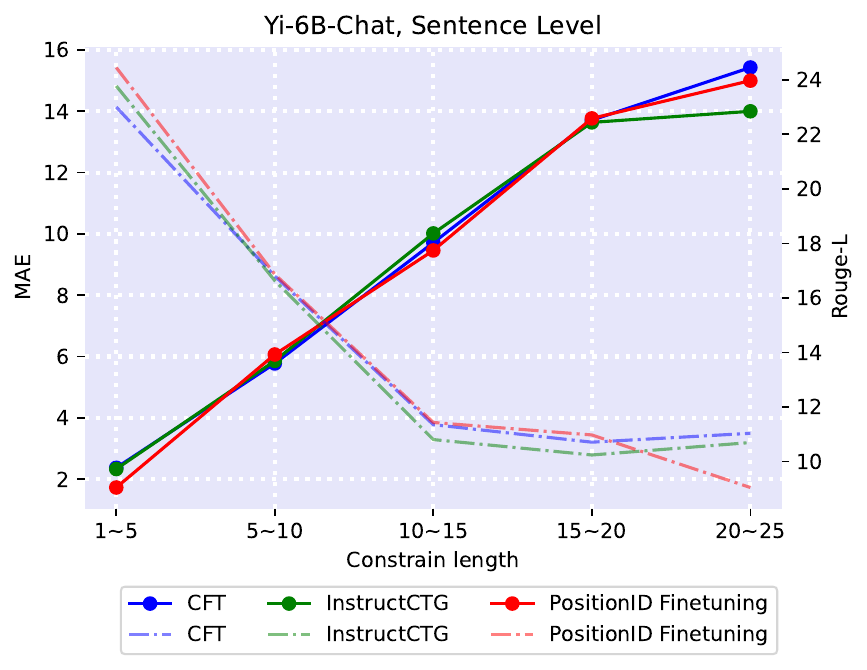}
    \caption{}
\end{subfigure}
\hfill
\begin{subfigure}[b]{0.32\textwidth}
    \includegraphics[width=\textwidth]{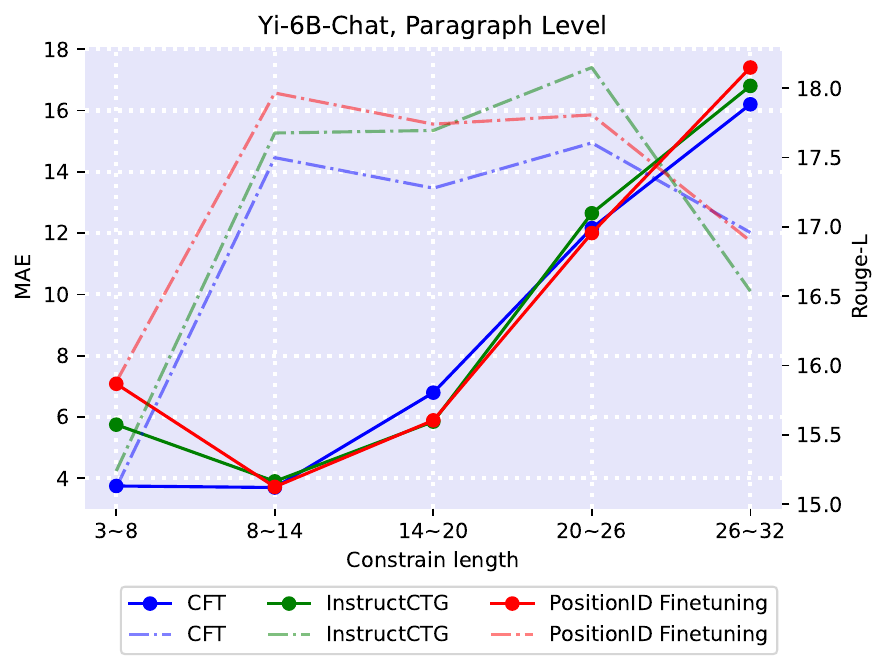}
    \caption{}
\end{subfigure}

\caption{Performance on LenCtrl-Bench with different constraint lengths. The solid line stands for the MAE, while the dotted line indicates the Rouge-L score. Figure a, b, and c showcase the performance of GPT-4 under PositionID prompting at different levels of granularity, while Figure d, e, and f display the results of Yi-6B-Chat after PositionID Fine-Tuning.}
\label{fig:evaluation_on_constrain_length}
\end{figure*}

\subsection{Main results}
We present our main results in Table \ref{tab:eval-prompts},  \ref{table:cp-bench-object} and \ref{table:cp-bench-subject}. Our key findings are summarized as follows:

\paragraph{PositionID Prompting is effective for Length Control.\label{sec:results-positionid-prompting}}

Compared to other length control prompting baselines, our method achieves the best performance across all levels. Additionally, regarding the Rouge-L metric, PositionID prompting achieves a score of 23.2, outperforming all other baselines. This indicates that PositionID prompting not only effectively controls the length of the generated responses but also maintains or even enhances the quality of the responses.

\paragraph{PositionID Fine-Tuning enhances model's positional awareness.}

Furthermore, in our length control fine-tuning experiments, we observe that PositionID fine-tuning consistently outperforms both CFT and InstructCTG in terms of MAE metrics across all levels. Notably, our method also achieves 19.8 in Rouge-L score, which is the closest to the zero-shot performance. This result suggests that our PositionID fine-tuning method not only enhances the accuracy of response length control but also preserves the quality of the generated responses compared to the baseline methods. At the paragraph level, PositionID fine-tuning does not yield the best MAE results compared to its initialization. We will discuss this result in \S\ref{sec:ablation_study}.

\paragraph{PositionID improves the functionality of the copy and paste tools}

For the copy-and-paste task, we evaluate from four perspectives in comparison to few-shot prompting. 
(1) The successful use of the tool: we notice a significant increase in the CP Success Rate of PositionID CP Prompting. Additionally, in terms of Consistency, our method achieves a higher win rate, implying that using the copy tool with position ids can better capture the key information of instruction. 
(2) Accuracy of the paste position: PositionID CP Prompting also takes the lead in PPL (8.4) and Clarity (33.7 winning rate). 
(3) Quality of the response content: PositionID CP Prompting achieved 18.4 and 55.8\% in Rouge-L and Accuracy respectively, surpassing the baseline. 
(4) Overall usage of the tools: PositionID scores 3.5 in tool-use proficiency, higher than 2.4 of few-shot.

\subsection{Ablation Studies}
\label{sec:ablation_study}

\paragraph{Generality of PositionID Granularities for PositionID Prompting.} 
We further investigate how different PositionID granularities influence the generation of responses. From Table \ref{table:ablation-on-prompt}, we observe that PositionID prompts at a specific level achieve the best MAE scores at its corresponding granularity, significantly outperforming PositionID prompts at other levels for the same granularity. For example, word-level PositionID Prompting performs the best at the word-level length constraint, but is sub-optimal for sentence-level length constraint. 
From Figure \ref{fig:total_results} in the Appendix \ref{appendix:mix_reuslts_constraint}, it's also apparent that each prompt at varying levels exhibits a significant decrease in the precision of length control across different granularities. This confirms that PositionID prompts at various levels can only enhance the model's ability to control length at the specific granularity, while adversely affecting at other granularities. 

\paragraph{Effectiveness of PositionID Prompting using Different Models.} 
We explore the performance of different models in length control tasks. We evaluate four models: GPT-4, Yi-6B-Chat, Mistral-7B-Instruct ~\citep{jiang2023mistral}, and Vicuna-7B ~\citep{vicuna2023}, under both zero-shot and PositionID Prompting settings. The results are presented in Figure \ref{fig:Different_Models}. Our findings reveal that: (1) GPT-4 outperforms all other models, achieving the highest scores in MAE at different levels, and demonstrating strong results in Rouge-L; (2) Among the open-source models, Mistral-7B-Instruct stands out as the best performer, particularly excelling at the word level, while Yi-6B-Chat ranks last; (3) However, the usage of PositionID Prompting leads to the most notable improvement for Yi-6B-Chat, reducing its MAE at the word level by 14.6. In contrast, Mistral-7B-Instruct and Vicuna-7B show a decrease in performance when employing PositionID Prompting.

\paragraph{Effectiveness of PositionID Prompting and PositionID Fine-Tuning under Different Length Constraints.} 
Additionally, we try to understand how the constraint length, as specified in the instructions, affects the model's output. We illustrate the results of PositionID Prompting with GPT, and PositionID Fine-Tuning with Yi-6B-Chat in Figure \ref{fig:evaluation_on_constrain_length}. We discover that: 

(1) At the word and sentence levels, MAE tends to rise with increasing constraint length, while the Rouge-L score exhibits an opposite trend. This indicates that the model faces challenges in generating longer texts: it not only loses precision in controlling the response length but also experiences a decline in the quality of the output. 
One possible reason for the increase in MAE could be the model's inability to accurately follow short length constraint requirement (1-57), which leads to cumulative errors when generating longer sentences. 

This could explain why PositionID did not perform optimally at the paragraph level for fune-tuning setting, as shown in Table \ref{tab:eval-prompts}: given that paragraphs are typically longer than words and sentences, the model's capacity to utilize position ids may decline.

(2) At the paragraph level, the lowest MAE is typically not observed at the beginning of the range (3-7 in Figure \ref{fig:evaluation_on_constrain_length} (c)), but rather occurs later (12-17 in Figure \ref{fig:evaluation_on_constrain_length} (c)). The Rouge-L score does not show a clear trend of increase or decrease. In fact, we find that the model tends to produce shorter paragraphs when required to generate more paragraphs. For example, in the results of GPT-4 with PositionID Prompting at the paragraph level, as the constraint length increases from 3-7 to 12-17, the average paragraph length decreases from 71.8 tokens to 57 tokens. As the constraint length increases further, the paragraph length remains approximately 56 tokens. This suggests that although more paragraphs are required to be generated, the interval between position ids decreases, potentially enhancing the model's sensitivity to text length and position id reasoning.

We refer the reader to Appendix \ref{appx:more-experiments} for more experiments including the more metrics for response quality evaluation, more results on the effect of constraint length on prompting, more models for PositionID Fine-Tuning, and ablation experiments on the mode mixing for PositionID Fine-Tuning. 

\section{Related Works}

Positional awareness of Large Language Models (LLMs) significantly impacts their performance~\citep{zhao2023survey, touvron2023llama, su2021roformer, abacus_embedding, contextual_position_encoding}. Techniques like Abacus Embeddings~\citep{abacus_embedding} and Contextual Position Encoding~\citep{contextual_position_encoding} improve tasks such as mathematical computation and selective copying but require model retraining and are limited to open-source LLMs. They also do not address length control or real-world copying and pasting.

Previous research on length-controlled text generation~\citep{qin2022cold, instructctg, sun2023evaluating} includes methods like InstructCTG~\citep{instructctg}, which uses fine-tuning with length constraints but is only applicable to open-source models. Our work introduces approaches suitable for both closed-source and open-source LLMs, focusing on length control and copying and pasting tasks.

\section{Conclusion}
In conclusion, our work introduces PositionID Prompting and PositionID Fine-Tuning to enhance large language models' positional awareness, enabling more accurate adherence to length control instructions without sacrificing response quality. Additionally, we propose PositionID CP Prompting to address the challenge of copy-paste operations. By explicitly marking the position of text units, PositionID CP Prompting allows models to accurately copy and paste specific spans of text. Our experiments on LenCtrl-Bench confirm improvements in length control, while tests on CP-Bench demonstrate the effectiveness of PositionID CP Prompting for copy and paste, highlighting its capability to maintain the integrity of the copied content and ensure precise placement during paste operations.

\section*{Limitations}

Despite the promising results of PositionID Prompting and PositionID Fine-Tuning, several limitations remain. First, these techniques require additional annotation and preprocessing steps, which may introduce complexity and overhead in practical applications. Second, the effectiveness of these methods is contingent on the granularity of PositionID assignments, which may vary across different tasks and use cases, potentially necessitating further fine-tuning and customization. Third, while our approaches improve positional awareness, they may still struggle with extremely fine-grained length control requirements, such as generating text with very long length requirements. Lastly, the generalizability of our techniques to closed-source models remains limited, as they may lead to double API costs due to the additional generation of position ids.

\section*{Ethics}

Our work is based on Large Language Models (LLMs), which can generate potentially harmful and unfaithful responses. Our proposed method aims to enhance the models' abilities for length control, and copy-paste. Therefore, our method presents little ethical issues. 
Our constructed data, LenCtrl-Bench and CP-Bench, don't involve any sensitive data.

\bibliography{custom}

\newpage
\appendix

\section{More Details about LenCtrl-Bench}
\label{appendix:lenctrl-bench-details}

\begin{figure}[h]
    \centering
    \includegraphics[width=\linewidth]{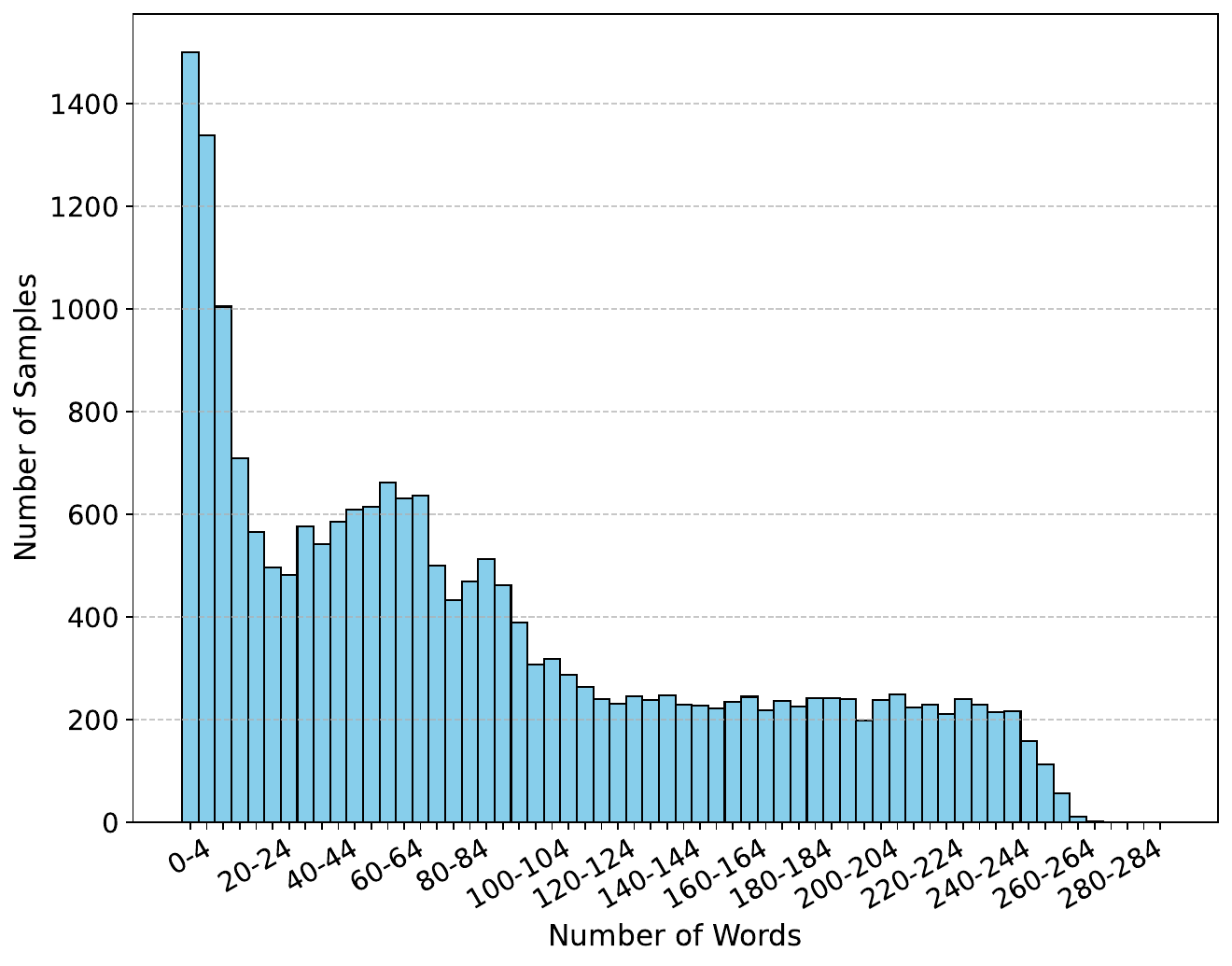}
    \caption{The distribution of the number of samples at word granularity in the training set.}
    \label{fig:train-word}
\end{figure}

\begin{figure}[h]
    \centering
    \includegraphics[width=\linewidth]{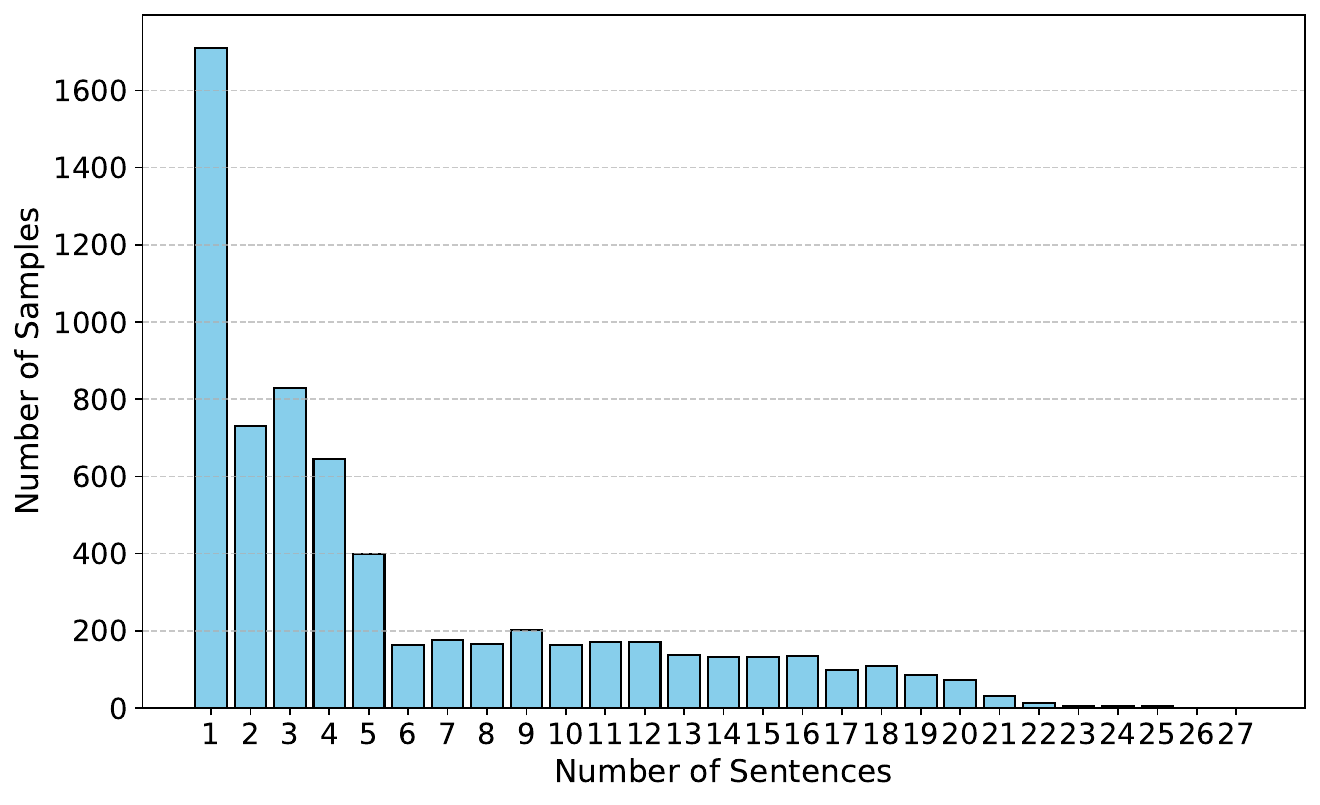}
    \caption{The distribution of the number of samples at sentence granularity in the training set.}
    \label{fig:train-sen}
\end{figure}

\begin{figure}[h]
    \centering
    \includegraphics[width=\linewidth]{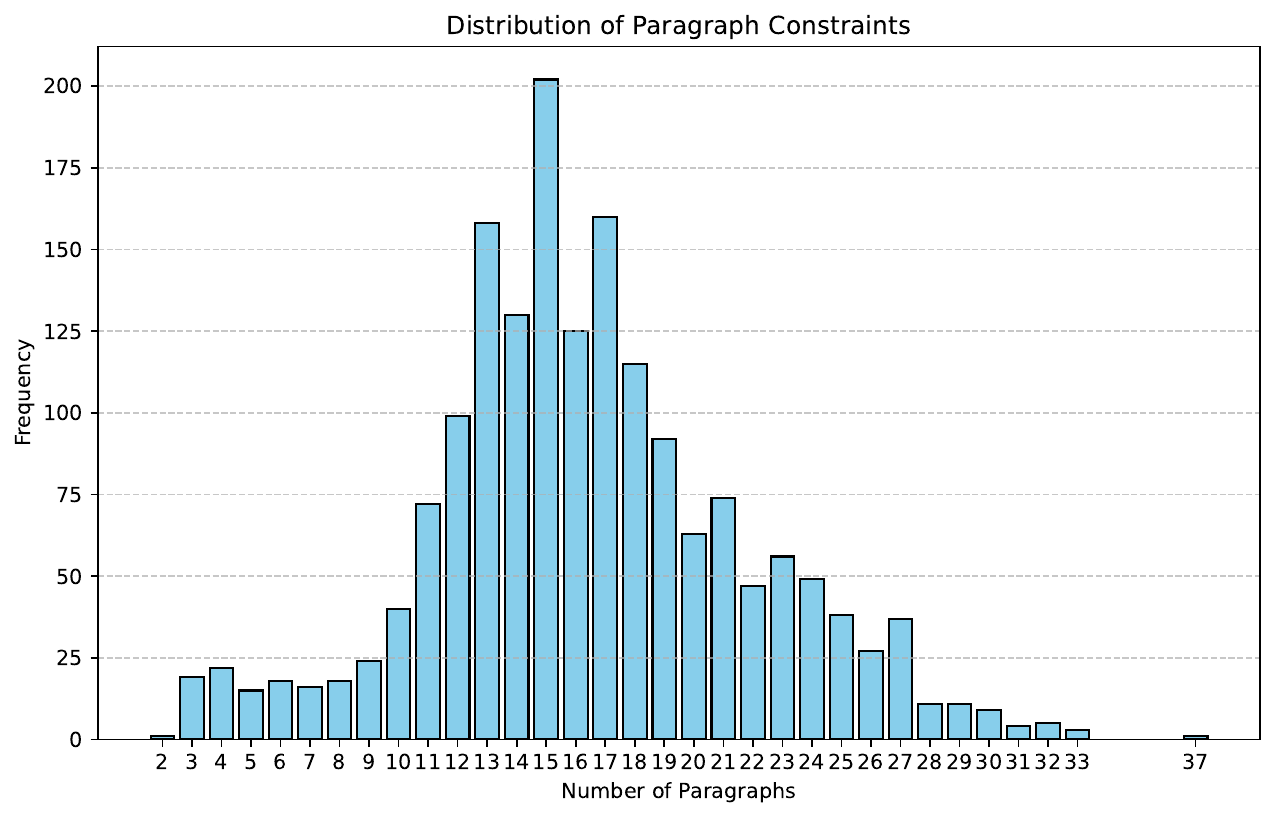}
    \caption{The distribution of the number of samples at paragraph granularity in the training set.}
    \label{fig:train-para}
\end{figure}

\begin{figure}[h]
    \centering
    \includegraphics[width=\linewidth]{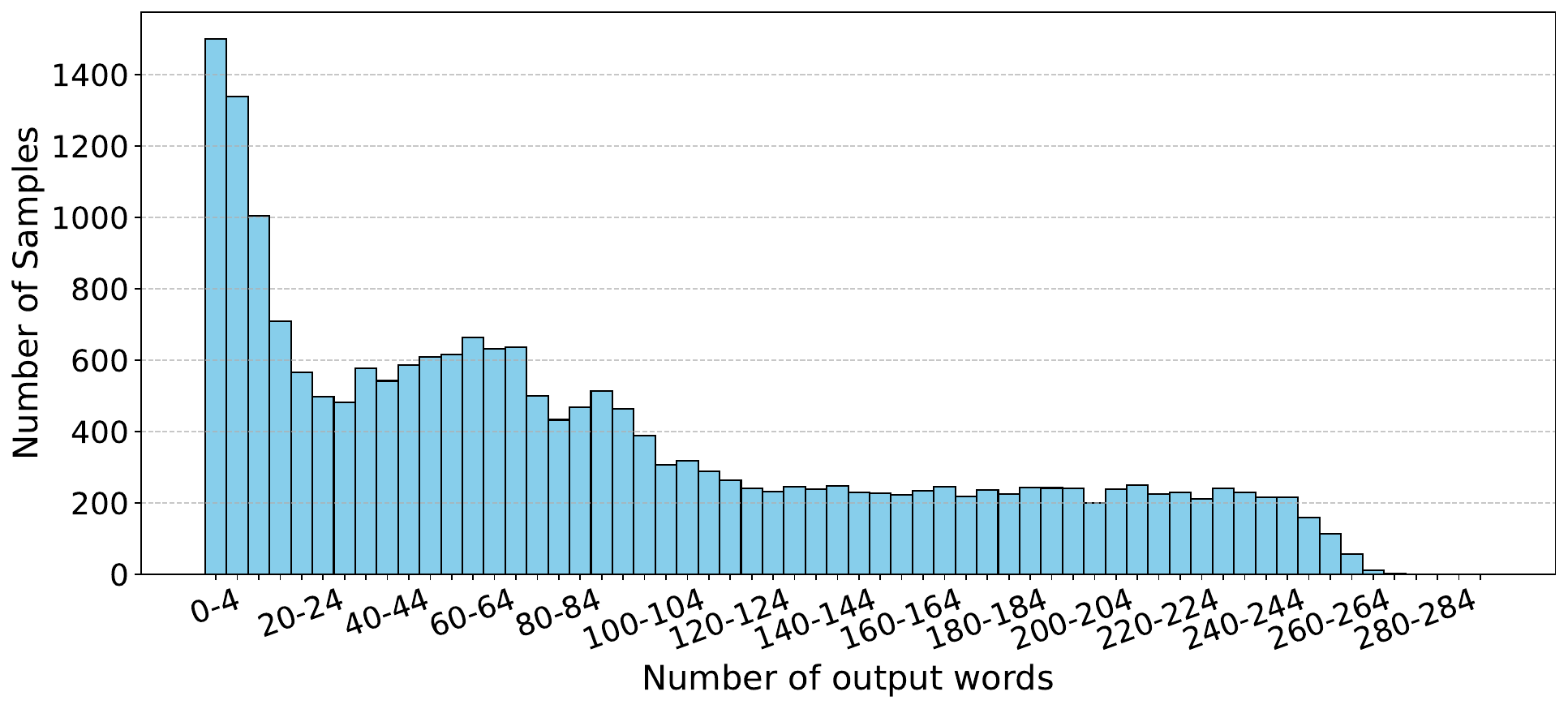}
    \caption{The length distribution of the training set with respect to the outputs.}
    \label{fig:train-output}
\end{figure}

The distribution of the number of words, sentences, and paragraphs in the training set is shown in Figures \ref{fig:train-word}, \ref{fig:train-sen} and \ref{fig:train-para}, respectively. The figures indicate that shorter texts are more common in the dataset, with the number of text samples decreasing as the length of the texts increases. In particular, the word count distribution shows a high concentration of samples in the 1 to 24 word range, especially for texts with 1 to 4 words. Similarly, the sentence count distribution is highest for texts with 1 sentence, and the paragraph count distribution peaks around 15 paragraphs. These data distributions suggest a comprehensive coverage of various text structures.
Similarly, the output length distribution, shown in Figure \ref{fig:train-output}, indicates that shorter texts are more common. The sample count is highest in the 1 to 24 word range, particularly for texts with 1 to 4 words. 

Overall, the dataset exhibits a balanced and representative sampling across different text lengths, ensuring a comprehensive coverage of various text structures.

\begin{figure}[h]
    \centering
    \includegraphics[width=\linewidth]{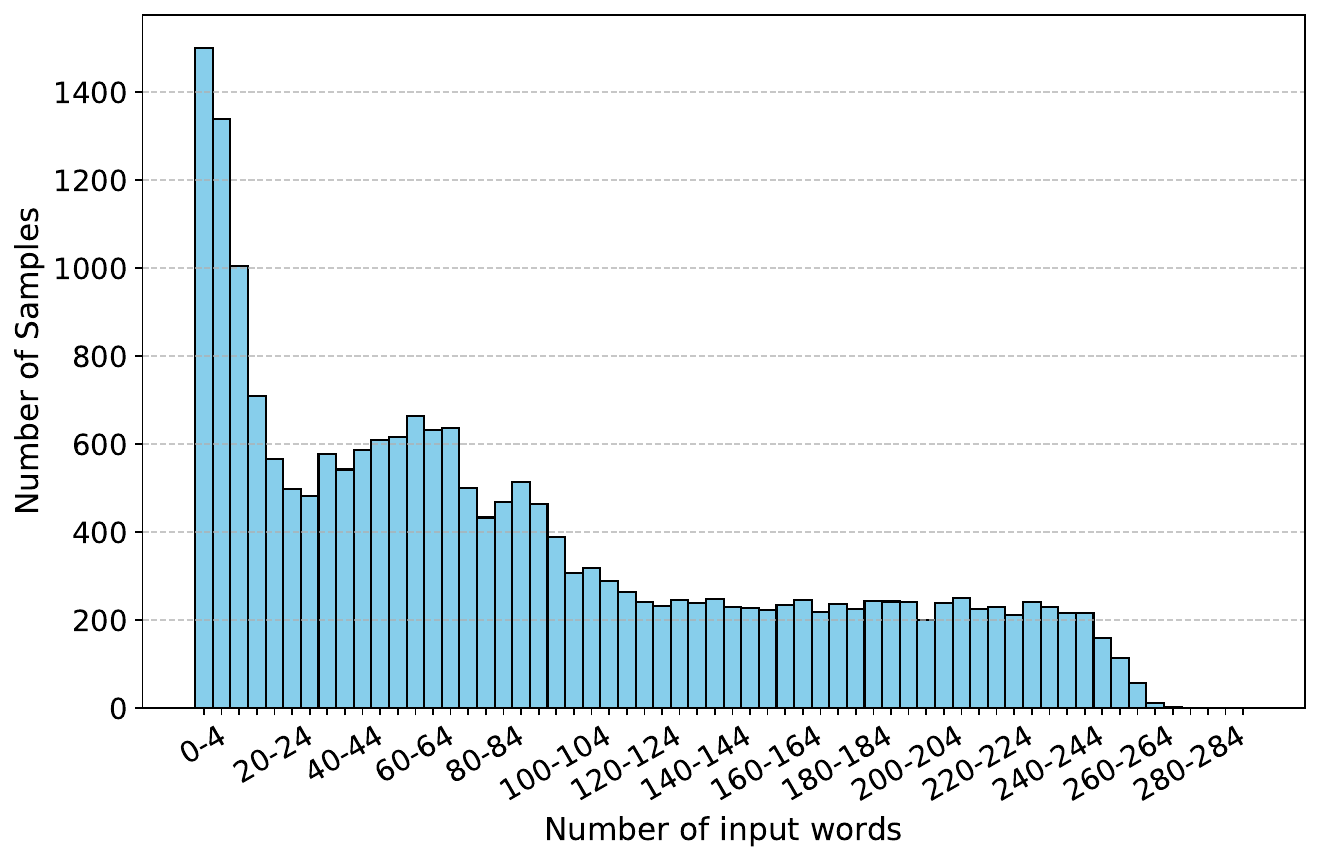}
    \caption{The length distribution of the training set with respect to the inputs.}
    \label{fig:train-input}
\end{figure}

\newpage


\begin{figure}[h]
    \centering
    \includegraphics[width=\linewidth]{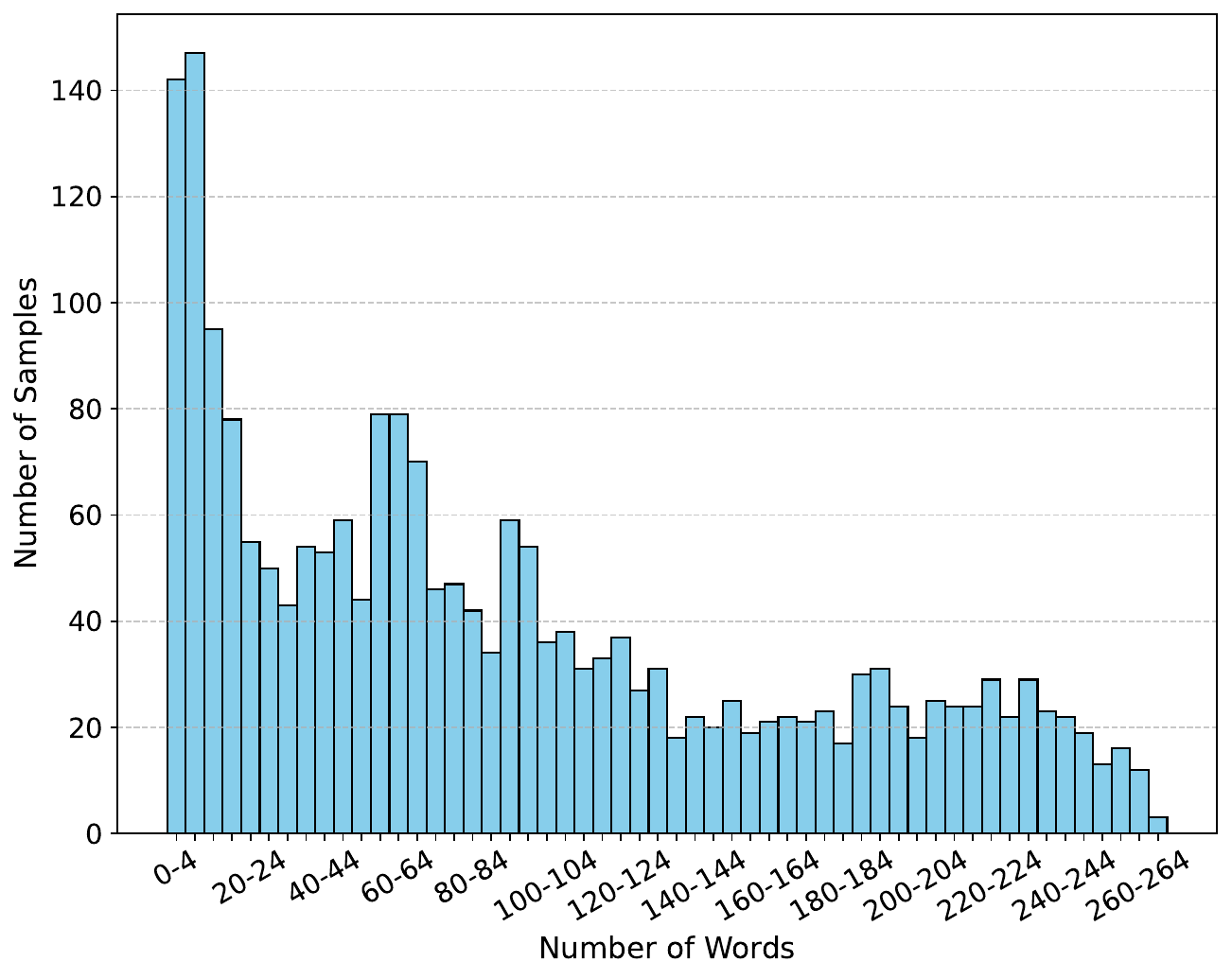}
    \caption{The distribution of the number of samples at word granularity in the test set.}
    \label{fig:test-word}
\end{figure}

\begin{figure}[h]
    \centering
    \includegraphics[width=\linewidth]{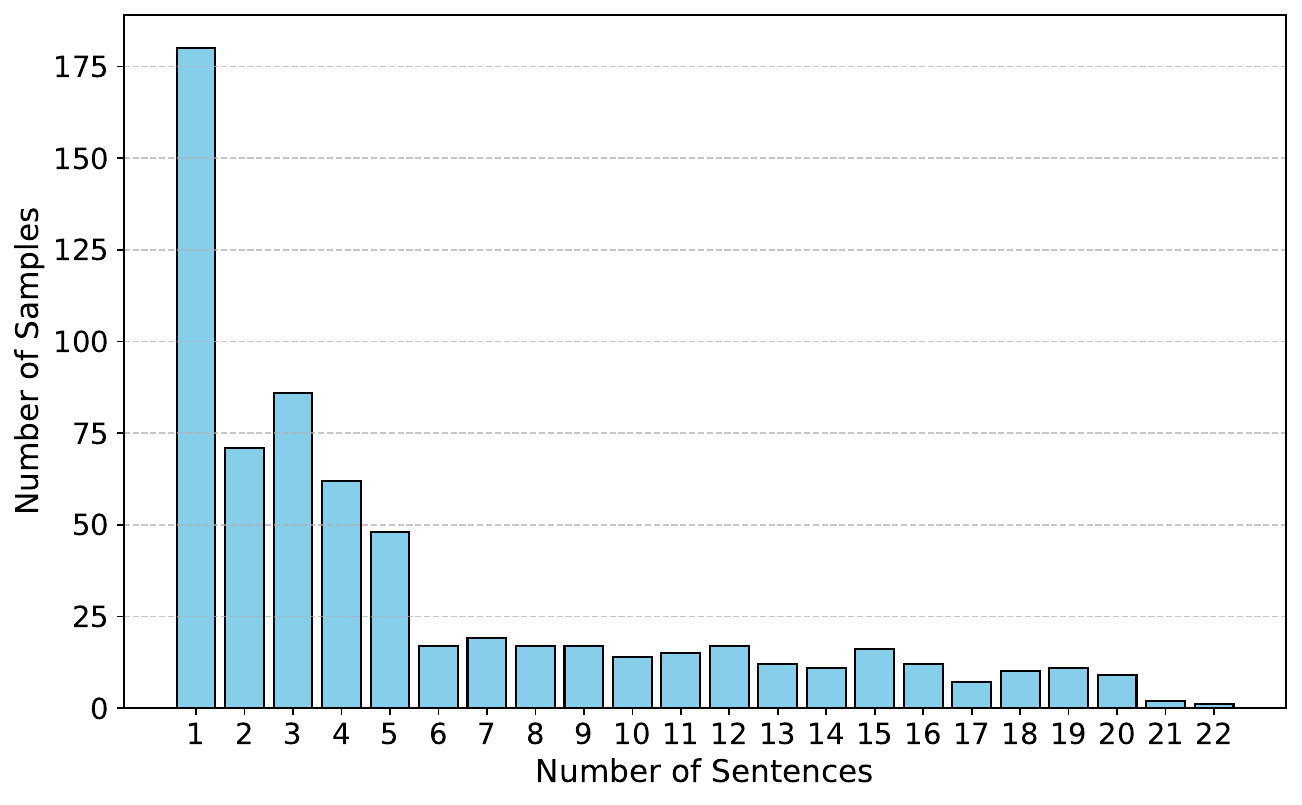}
    \caption{The distribution of the number of samples at sentence granularity in the test set.}
    \label{fig:test-sen}
\end{figure}

\begin{figure}[h]
    \centering
    \includegraphics[width=\linewidth]{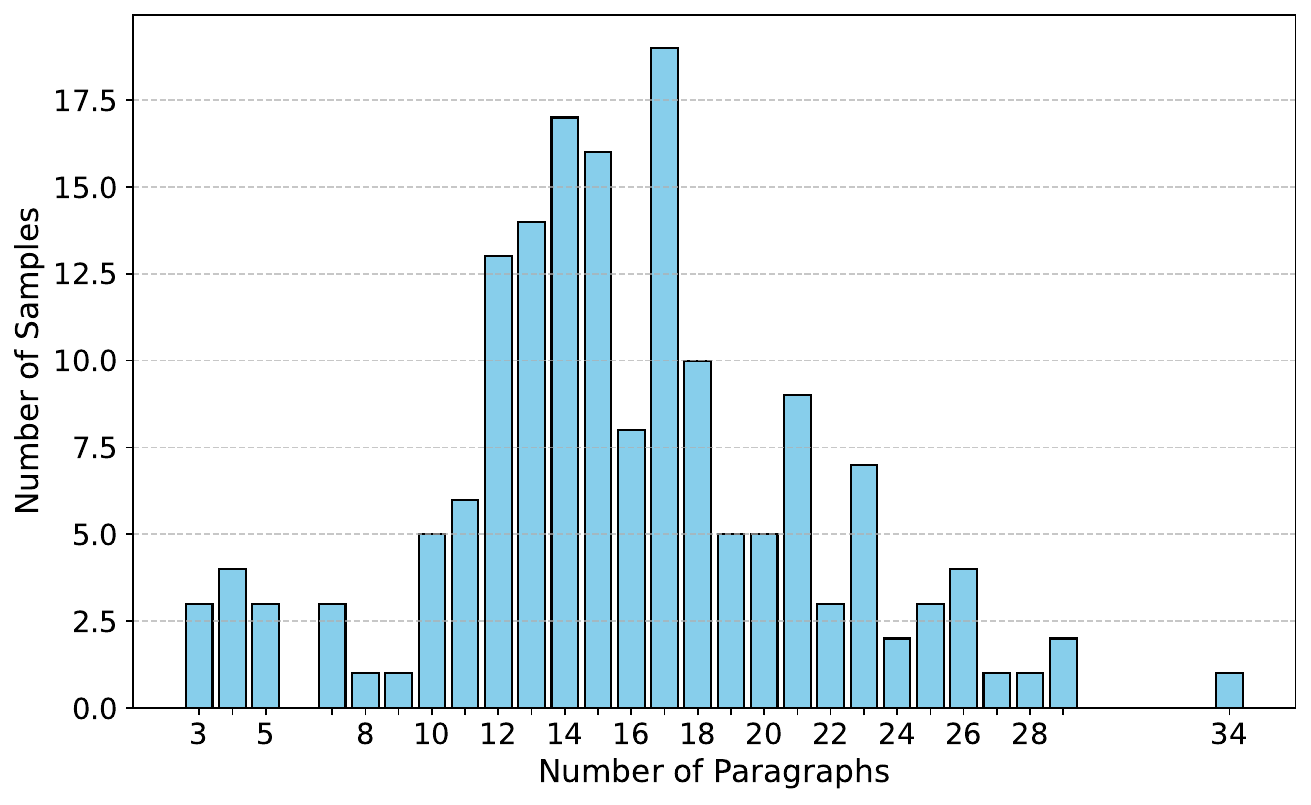}
    \caption{The distribution of the number of samples at paragraph granularity in the test set.}
    \label{fig:test-para}
\end{figure}

\begin{figure}[h]
    \centering
    \includegraphics[width=\linewidth]{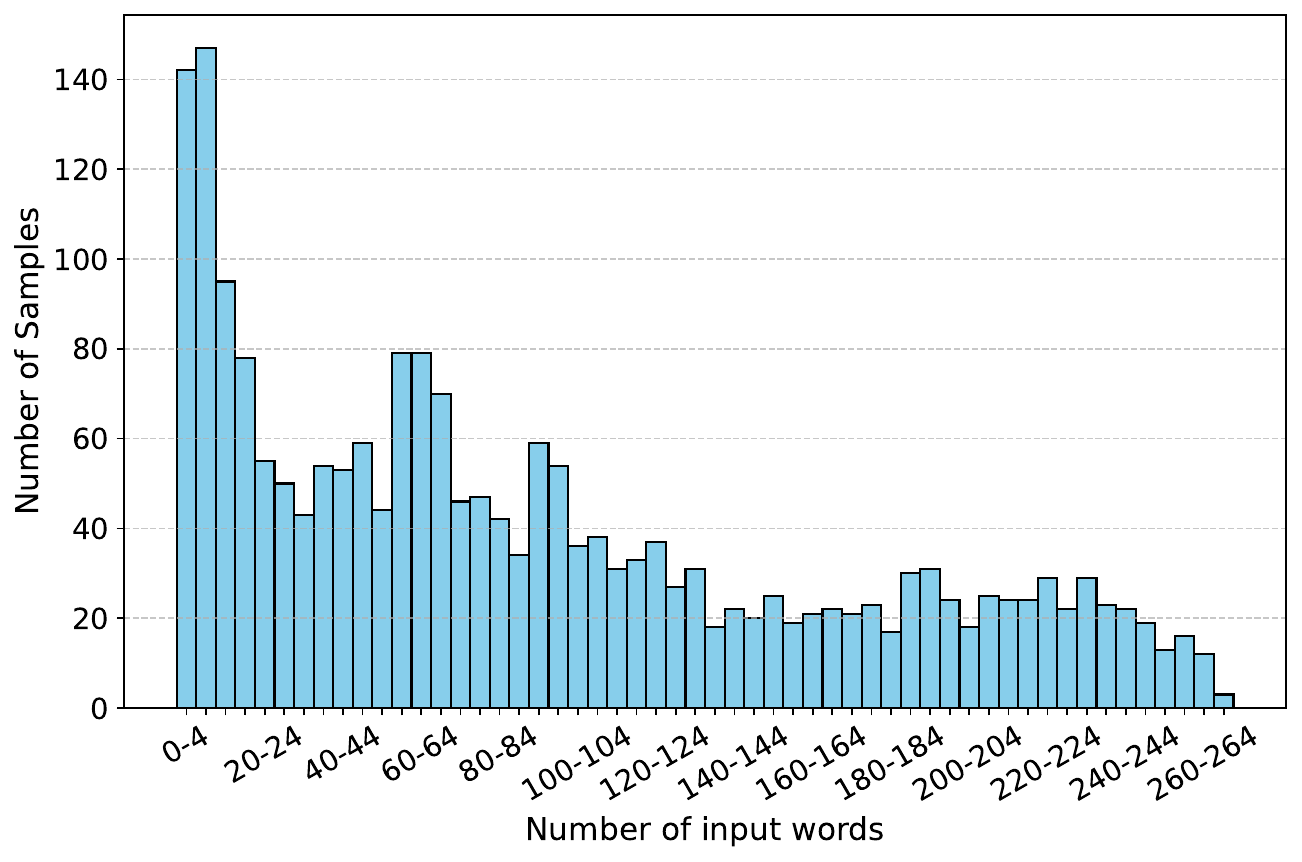}
    \caption{The length distribution of the test set with respect to the inputs.}
    \label{fig:test-input}
\end{figure}

\begin{figure}[h]
    \centering
    \includegraphics[width=\linewidth]{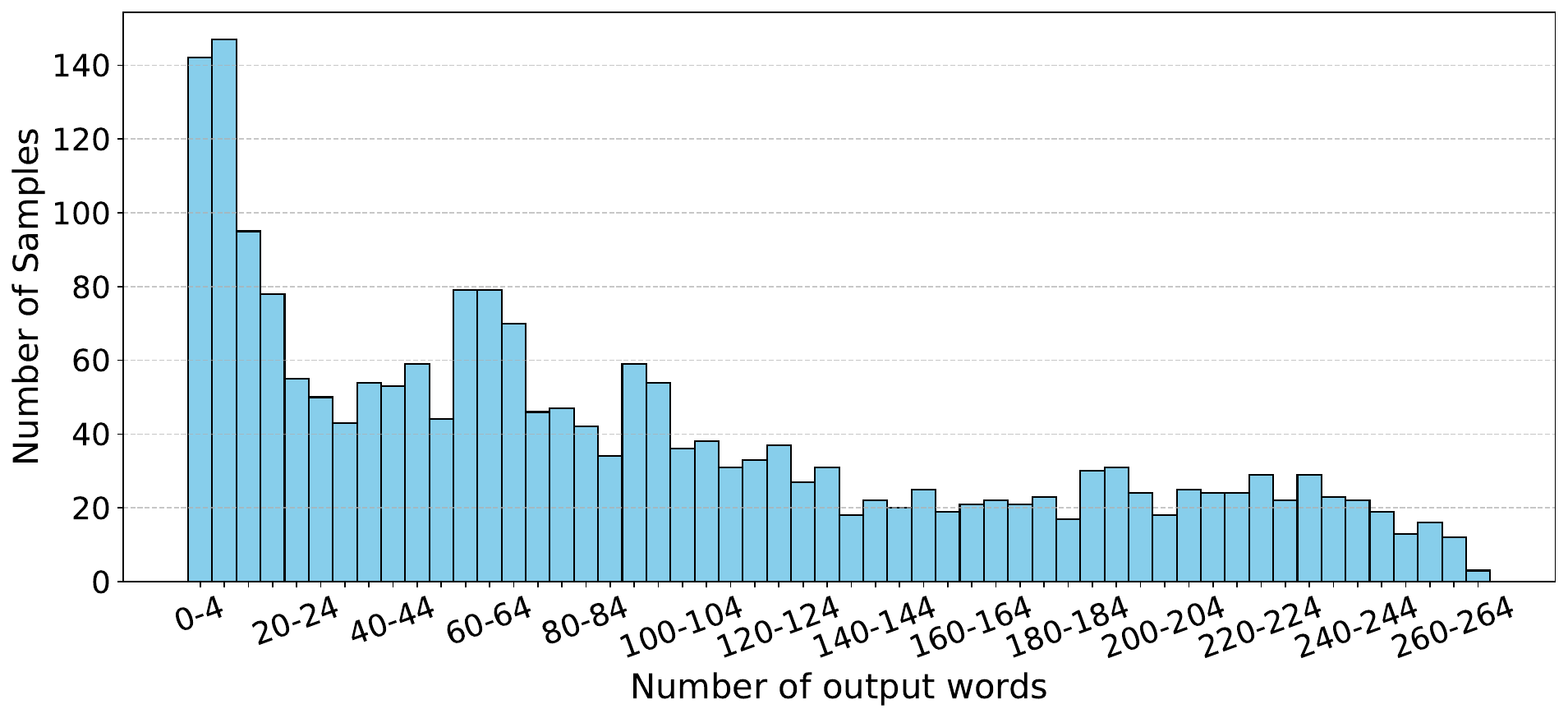}
    \caption{The length distribution of the test set with respect to the outputs.}
    \label{fig:test-output}
\end{figure}

The distributions of various text lengths in the test set are shown in Figures \ref{fig:test-word}, \ref{fig:test-sen}, \ref{fig:test-para}, \ref{fig:test-input}, and \ref{fig:test-output}.

Figure \ref{fig:test-word} displays the distribution of text samples with varying word numbers, indicating that shorter texts are more common. Figure \ref{fig:test-sen} shows the distribution of text samples with varying sentence numbers, again highlighting the prevalence of shorter texts.
The paragraph count distribution is shown in Figure \ref{fig:test-para}, which follows a normal distribution trend centered around 14 to 18 paragraphs.
Figures \ref{fig:test-input}, and \ref{fig:test-output} present the input and output word length distributions, respectively, both showing that shorter texts are more frequent.

Overall, the test set exhibits a balanced distribution of text samples across varying lengths, ensuring comprehensive coverage of different text structures, which is crucial for robust data analysis and model testing.

\newpage
\section{More Details about CP-Bench}
\label{appendix:cpbench-detail}

\begin{figure}[h]
    \centering
    \includegraphics[width=\linewidth]{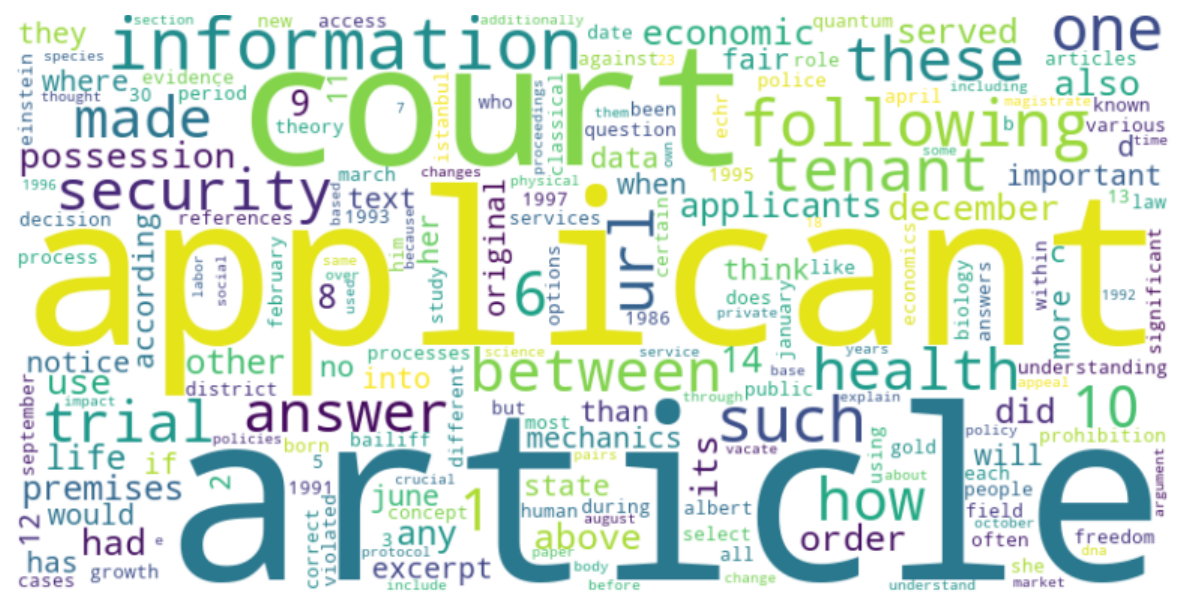}
    \caption{The Word Cloud of CP-Bench.}
    \label{fig:word cloud}
\end{figure}

As shown in Table \ref{tab:cp-patterns}, we have 6 copy-paste patterns constructed from the open-source dataset: Option Selection, Law Article Statement, Policy Statement , URL, Quotation, and General Text Spans, and they have a proportion of 12.1, 11.0, 10.4, 12.6, 6.6, 11.0 of the total dataset. We also have 2 copy-paste patterns synthesized by GPT-4o, and they have a proportion of 19.2, 17.0 of the total dataset. The word cloud diagram of the CP-Bench is shown in Figure \ref{fig:word cloud}.

As for the construction details of CP-Bench, the subsets for patterns such as option selection, quotation, and URL are created by extracting text spans using regular expression (regex) patterns, as listed in Table \ref{tab:regex-patterns}. For patterns like law article statements, policy statements, and general text spans, we directly extract repeated sequences of consecutive sentences. Subsets in patterns like terminology reiteration and note-taking are synthesized by prompting GPT with several human-written examples. To ensure quality, all final samples in our CP-Bench are filtered by human annotators.

\newpage

\begin{table*}[h]
    \centering
    \small
    \begin{tabular}{>{\centering\arraybackslash}m{3cm}|>{\centering\arraybackslash}m{4cm}
    |>{\centering\arraybackslash}m{1cm}
    |>{\centering\arraybackslash}m{5cm}}
    \toprule
        \textbf{Pattern} & \textbf{Description} & \textbf{Ratio} & \textbf{Construction}
        \\\midrule
        \textbf{Option Selection} & Copy the content of an option in multi-choice questions & 12.1 & Constructed from OpenHermes 2.0~\citep{OpenHermes} \\ \midrule
        \textbf{Law Article Statement} & Quote the law article that appears above & 11.0 & Constructed from LexGLUE~\citep{chalkidis-etal-2021-lexglue} \\ \midrule
        \textbf{Policy Statement} & Cite the policy statement to answer the questions asked & 10.4 & Constructed from PrivacyQA~\citep{ravichander2019question}\\ \midrule
        \textbf{Terminology Reiteration} & Reiterate a terminology & 19.2 & GPT Synthesis \\ \midrule
        \textbf{Note-Taking} & Extract some keywords, key phrases, or key sentences from the provided context & 17.0 & GPT Synthesis \\ \midrule
        \textbf{URL} & Copy a long URL & 12.6 & Constructed from OpenHermes 2.0~\citep{OpenHermes}\\ \midrule
        \textbf{Quotation} & Quote a sentence enclosed in either single or double quotes & 6.6 & Constructed from OpenHermes 2.0~\citep{OpenHermes}\\ \midrule
        \textbf{General Text Spans} & Repeat key text spans & 11.0 & Constructed from SQuAD 2.0~\citep{rajpurkar2018squad2}, NewsQA~\citep{trischler2016newsqa} and ReClor~\citep{yu2020reclor}\\ 
    \bottomrule
    \end{tabular}
    \caption{Eight copy-paste patterns.}
    \label{tab:cp-patterns}
\end{table*}

\begin{table*}[ht]
    \centering
    \small
    \begin{tabular}{c|c}
    \toprule
    \textbf{CP Patterns} & \textbf{Regex Patterns} \\
    \midrule
    \textbf{Option Selection} & \verb|r'(?<=:)(.*?\.(?!\d))'| \\ \midrule
    \textbf{Quotation} & \verb|r'"\'["\']'| \\ \midrule
    \textbf{URL} & \verb|r'\b(?:https?:\/\/)?(?:www\.)?[a-zA-Z0-9.-]+\.[a-zA-Z]{2,}(?:\/[^\s]*)?\/?'| \\
    \bottomrule
    \end{tabular}
    \caption{Regex patterns for extracting copy-paste spans for CP-Bench.}
    \label{tab:regex-patterns}
\end{table*}

\section{Implementation Details}
\label{appendix:implementation_details}

\paragraph{Prompt Templates.} 
The prompt templates used for PositionID Prompting, PositionID Fine-Tuning, PositionID CP Prompting, and Few-Shot CP Prompting are shown in Boxes \ref{appx:prompt-positionid}.


\onecolumn
\phantomsection
\label{appx:prompt-positionid}
\begin{tcolorbox}[colback=white!95!gray,colframe=gray!50!black,rounded corners,label={prompt-dot}, title={Prompt Template (PositionID Prompting) (Word-Level)}]
\begin{lstlisting}[breaklines=true, xleftmargin=0pt, breakindent=0pt, columns=fullflexible, mathescape]
System prompt at word level:
```
###Instruction
You are a helpful AI assistant.
Format your response by appending a sequential ID number to each token.
Make sure the sentence structure remains intact and clear, and each word should be distinctly numbered.
You need to follow the input instruction and stop right at the position id equal to the required word number.
Remember, punctuation doesn't count for numbering

Demonstrations:
###Input
Find the synonyms for the following words and write them down in one sentence. Please be succinct: limit your response to 7 words.
###Response
Meticulous 1 - attentive 2 ; elusive 3 - evasive 4 ; supple 5 - 6 flexible 7 . 

{more demonstrations}
```
\end{lstlisting}
\end{tcolorbox}

\phantomsection
\label{appx:prompt-positionid}
\begin{tcolorbox}[colback=white!95!gray,colframe=gray!50!black,rounded corners,label={prompt-dot}, title={Prompt Template (PositionID Prompting) (Sentence-Level)}]
\begin{lstlisting}[breaklines=true, xleftmargin=0pt, breakindent=0pt, columns=fullflexible, mathescape]
System prompt at sentence level:
```
###Instruction
You are a helpful AI assistant.
Format your response by appending a sequential ID number to each sentence.
Make sure the sentence structure remains intact and clear, and each sentence is distinctly numbered.
You need to follow the input instruction and stop right at the sentence ID equal to the required sentence number.
Remember, punctuation doesn't count for numbering

Demonstrations:
###Input
Identify what kind of bias is present in the given statement. With a response length of 3 sentences.
###Response
This statement shows a bias towards bosses in general. <1> It implies that there are certain characteristics associated with bosses that make them difficult to work with, regardless of the individual boss. <2> This statement assumes an all-or-nothing attitude and perpetuates a negative stereotype thatbosses in general are difficult to work with. <3>

{more demonstrations}
```
\end{lstlisting}
\end{tcolorbox}

\phantomsection
\label{appx:prompt-positionid}
\begin{tcolorbox}[colback=white!95!gray,colframe=gray!50!black,rounded corners,label={prompt-dot}, title={Prompt Template (PositionID Prompting) (Paragraph-Level)}]
\begin{lstlisting}[breaklines=true, xleftmargin=0pt, breakindent=0pt, columns=fullflexible, mathescape]
System prompt at paragraph level:
```
###Instruction
You are a helpful AI assistant.
Format your response by appending a sequential ID number to each paragraph.
Make sure the sentence structure remains intact and clear, and each paragraph is distinctly numbered.
You need to follow the input instruction and stop right at the paragraph ID equal to the required paragraph number.
Remember, each paragraph is separated by two line breaks.

Demonstrations:
###Input
Generate two points that could be used to extend the following statement Limit your narrative to 2 paragraphs.
###Response
- It has also become a key tool for collaboration and communication .

 <1> - It has provided a platform for the sharing of ideas and greater access to global markets .

 <2>

{more demonstrations}
\end{lstlisting}
\end{tcolorbox}

\phantomsection
\label{appx:prompt-positionid-ft}
\begin{tcolorbox}[colback=white!95!gray,colframe=gray!50!black,rounded corners,label={prompt-dot}, title={Prompt Template (PositionID Fine-Tuning)}]
\begin{lstlisting}[breaklines=true, xleftmargin=0pt, breakindent=0pt, columns=fullflexible, mathescape]
System Prompt in PositionID Mode:
```
You are an assistant replying in the PositionID mode. Please strictly follow the length constraint to generate your response. You can generate your response while continuously counting the response length to facilitate the length control.
 
The PositionID indicates the position of each word (or sentence or paragraph) in the text. 
Specifically, the first word in the text is assigned the PositionID `1`, the second word is assigned the PositionID `2`, and so on. 
The first sentence in the text is assigned the PositionID `<1>`, the second sentence is assigned the PositionID `<2>`, and so on. 
The first paragraph in the text is assigned the PositionID `<1>`, the second paragraph is assigned the PositionID `<2>`, and so on.

Examples: 
If the length constraint is 10 words, you should generate a response like this: "This 1 is 2 an 3 example 4 of 5 a 6 response 7 with 8 ten 9 words 10."
If the length constraint is 3 sentences, you should generate a response like this: "This is the first sentence. <1> This is the second sentence. <2> This is the third sentence. <3>"
If the length constraint is 2 paragraphs, you should generate a response like this: "This is the first paragraph. <1> \n\n This is the second paragraph. <2>"
```

System Prompt in Normal Mode:
```
You are an assistant replying in the normal mode. Please strictly follow the length constraint to generate your response.

For example, if the length constraint is 10 words, you should generate a response like this: ``This is an example of a response with ten words.''
If the length constraint is 3 sentences, you should generate a response like this: ``This is the first sentence. This is the second sentence. This is the third sentence.''
If the length constraint is 2 paragraphs, you should generate a response like this: ``This is the first paragraph. \n\n This is the second paragraph.''
```
\end{lstlisting}
\end{tcolorbox}

\phantomsection
\label{appx:prompt-fewshot-cp}
\begin{tcolorbox}[colback=white!95!gray,colframe=gray!50!black,rounded corners,label={prompt-dot}, title={Prompt Template (PositionID CP Prompting) (Pre-Generation and Paste Tool Calling)}]
\begin{lstlisting}[breaklines=true, xleftmargin=0pt, breakindent=0pt, columns=fullflexible, mathescape]
System prompt, during pre-generation and paste tool calling phase:
```
You are a helpful AI assistant.

You are equipped with two tools:
1. **Copy Tool**: If you need to repeat some spans in your response, you have the ability to invoke the Copy Tool. This tool can copy the spans that you want to repeat in your response into the clipboard.
2. **Paste Tool**: This tool is associated with the copy tool. You can use the Paste Tool when you want to repeat a copied span in the clipboard and insert it in the current position.

To name a few, here are some scenarios where you can use the Copy Tool and the Paste Tool:
1. **Option Selection**: When you need to copy the content of an option in multi-choice questions, you can use the Copy Tool and then the Paste Tool to insert the option content into your response.
2. **Law Article Statement**: When you need to quote a law article, you can use the Copy Tool and then the Paste Tool to insert the law article into your response. 
3. **Terminology Reiteration**: When you need to reiterate a terminology, you can use the Copy Tool and then the Paste Tool to insert the terminology into your response.
4. **Note-Taking**: When you find it necessary to extract some keywords, key phrases, or key sentences from the provided context, you can use the Copy Tool and then the Paste Tool to put down the notes.
5. **URL**: When you find a long URL in the context, it's better to copy it using the Copy Tool and then the Paste Tool to insert it in your response.
6. **Quotation**: When you need to quote a sentence enclosed in either single or double quotes, you can use the Copy Tool and then the Paste Tool to insert the quoted sentence into your response.
7. **Policy Statement**: When you need to quote a policy statement, you can use the Copy Tool and then the Paste Tool to insert the policy statement into your response.
8. **General Text Spans**: Of course, you can use the Copy Tool and the Paste Tool to repeat any text spans that you find necessary to repeat in your response.

Here are the APIs for the Copy Tool and the Paste Tool:
1. **Copy Tool**: `<COPY>[tag="tag1"][description="The description of this copied span."][start=12][end=15]</COPY>.`
    This tool is a function with four parameters: `tag`, `description`, `start`, and `end`.
    In this API, you need to specify the tag, the description of the copied span, and the start and end word positions (including punctuations and the end) of the span that you want to copy.
    You must wrap all the parameters with `<COPY>` and `</COPY>`.
2. **Paste Tool**: `<PASTE>[tag="tag1"]</PASTE>.`
    This tool is a function with one parameter: `tag`, which is strictly the same as the `tag` designed in the previous calling of the Copy Tool.
    Note that there may be multiple calls of the Copy Tool in the previous text, you must use the accurate `tag` to paste the corresponding copied span.
\end{lstlisting}
\end{tcolorbox}

\phantomsection
\label{appx:prompt-fewshot-cp}
\begin{tcolorbox}[colback=white!95!gray,colframe=gray!50!black,rounded corners,label={prompt-dot}, title={Prompt Template (PositionID CP Prompting) (Pre-Generation and Paste Tool Calling) (Continued)}]
\begin{lstlisting}[breaklines=true, xleftmargin=0pt, breakindent=0pt, columns=fullflexible, mathescape]
To help you determine the `start` and the `end` parameters more accurately, the position ids are assigned when you start calling the Copy Tool by generating the `<COPY>` token. 
The position id indicates the position of each word including the punctuations in the text. 
Specifically, the first word in the text is assigned the position id `1`, the second word is assigned the position id `2`, and so on. 
For example, the sentence "The capital of France is Paris." has the position ids as follows: "The [1] capital [2] of [3] France [4] is [5] Paris [6]. [7]".
If you want to copy the word "Paris" in the sentence, you can use the Copy Tool with the `start` and `end` parameters as `[start=6][end=6]`.

Here are some examples to illustrate the usage of the Copy Tool and the Paste Tool (without showing position ids):

---
**Example 1:**

User: 
What is the capital of France?
a) Berlin  
b) Madrid  
c) Paris  
d) Rome  

Assistant: 
The capital of France is Paris <COPY>[tag="option_c"][description="Option C: Paris"][start=14][end=16]</COPY>. Therefore, the correct answer is <PASTE>[tag="option_c"]</PASTE>.
---
{more demonstrations}

Very Important: 
1. Don't copy the same span over and over again.
2. You must use the Copy Tool and Paste Tool in your response.
3. All the copied spans must be pasted.
4. There may be multiple different Copy Tool Calls, you must accurately paste the right one.
5. The position ids are just used for your reference. Don't show them in your response. And you must not count the position ids when determining the `start` and `end` parameters.
6. For the usage of position ids, remind the exmaple: "The capital of France is Paris." has the position ids as follows: "The [1] capital [2] of [3] France [4] is [5] Paris [6]. [7]". If you want to copy the word "Paris" in the sentence, you can use the Copy Tool with the `start` and `end` parameters as `[start=6][end=6]`.
```
\end{lstlisting}
\end{tcolorbox}

\phantomsection
\label{appx:prompt-fewshot-cp}
\begin{tcolorbox}[colback=white!95!gray,colframe=gray!50!black,rounded corners,label={prompt-dot}, title={Prompt Template (PositionID CP Prompting) (Copy Tool Calling)}]
\begin{lstlisting}[breaklines=true, xleftmargin=0pt, breakindent=0pt, columns=fullflexible, mathescape]
System prompt, during copy tool calling phase:
```
You are a helpful AI assistant.

You are equipped with one tool:
1. **Copy Tool**: If you need to repeat some spans in your response, you have the ability to invoke the Copy Tool. This tool can copy the spans that you want to repeat in your response into the clipboard.
Your target is to continue generating the content of the Copy tool

To name a few, here are some scenarios where you can use the Copy Tool and the Paste Tool:
1. **Option Selection**: When you need to copy the content of an option in multi-choice questions, you can use the Copy Tool and then the Paste Tool to insert the option content into your response.
2. **Law Article Statement**: When you need to quote a law article, you can use the Copy Tool and then the Paste Tool to insert the law article into your response. 
3. **Terminology Reiteration**: When you need to reiterate a terminology, you can use the Copy Tool and then the Paste Tool to insert the terminology into your response.
4. **Note-Taking**: When you find it necessary to extract some keywords, key phrases, or key sentences from the provided context, you can use the Copy Tool and then the Paste Tool to put down the notes.
5. **URL**: When you find a long URL in the context, it's better to copy it using the Copy Tool and then the Paste Tool to insert it in your response.
6. **Quotation**: When you need to quote a sentence enclosed in either single or double quotes, you can use the Copy Tool and then the Paste Tool to insert the quoted sentence into your response.
7. **Policy Statement**: When you need to quote a policy statement, you can use the Copy Tool and then the Paste Tool to insert the policy statement into your response.
8. **General Text Spans**: Of course, you can use the Copy Tool and the Paste Tool to repeat any text spans that you find necessary to repeat in your response.

Here are the APIs for the Copy Tool:
1. **Copy Tool**: `<COPY>[tag="tag1"][description="The description of this copied span."][start=12][end=15]</COPY>.`
    This tool is a function with four parameters: `tag`, `description`, `start`, and `end`.
    In this API, you need to specify the tag, the description of the copied span, and the start and end word positions (including punctuations and the end) of the span that you want to copy.
    You must wrap all the parameters with `<COPY>` and `</COPY>`.

To help you determine the `start` and the `end` parameters more accurately, the position ids are assigned when you start calling the Copy Tool by generating the `<COPY>` token. 
The position id indicates the position of each word including the punctuations in the text. 
Specifically, the first word in the text is assigned the position id `1`, the second word is assigned the position id `2`, and so on. 
For example, the sentence "The capital of France is Paris." has the position ids as follows: "The [1] capital [2] of [3] France [4] is [5] Paris [6]. [7]".
If you want to copy the word "Paris" in the sentence, you can use the Copy Tool with the `start` and `end` parameters as `[start=6][end=6]`.
\end{lstlisting}
\end{tcolorbox}

\phantomsection
\label{appx:prompt-fewshot-cp}
\begin{tcolorbox}[colback=white!95!gray,colframe=gray!50!black,rounded corners,label={prompt-dot}, title={Prompt Template (PositionID CP Prompting) (Copy Tool Calling) (Continued)}]
\begin{lstlisting}[breaklines=true, xleftmargin=0pt, breakindent=0pt, columns=fullflexible, mathescape]
Here are some examples to illustrate the usage of the Copy Tool:

---
**Example 1:**

User: 
What [1] is [2] the [3] capital [4] of [5] France [6] ? [7] a [8] ) [9] Berlin [10] b [11] ) [12] Madrid [13] c [14] ) [15] Paris [16] d [17] ) [18] Rome [19] The [20] capital [21] of [22] France [23] is [24] <COPY>

Assistant: 
[tag="option_c"][description="Option C: Paris"][start=14][end=16]</COPY>
---
{more demonstrations}

Very Important: 
1. Don't copy the same span over and over again.
2. You must use the Copy Tool in your response.
3. The position ids are just used for your reference. Don't show them in your response. And you must not count the position ids when determining the `start` and `end` parameters.
4. For the usage of position ids, remind the example: "The capital of France is Paris." has the position ids as follows: "The [1] capital [2] of [3] France [4] is [5] Paris [6]. [7]". If you want to copy the word "Paris" in the sentence, you can use the Copy Tool with the `start` and `end` parameters as `[start=6][end=6]`.
```
\end{lstlisting}
\end{tcolorbox}

\newpage
\phantomsection
\label{appx:prompt-positionid-cp}
\begin{tcolorbox}[colback=white!95!gray,colframe=gray!50!black,rounded corners,label={prompt-dot}, title={Prompt Template (Few-Shot CP Prompting) (Pre-Generation and Paste Tool Calling)}]
\begin{lstlisting}[breaklines=true, xleftmargin=0pt, breakindent=0pt, columns=fullflexible, mathescape]
System prompt, during pre-generation and paste tool calling phase:
```
You are a helpful AI assistant.

You are equipped with two tools:
1. **Copy Tool**: If you need to repeat some spans in your response, you have the ability to invoke the Copy Tool. This tool can copy the spans that you want to repeat in your response into the clipboard.
2. **Paste Tool**: This tool is associated with the copy tool. You can use the Paste Tool when you want to repeat a copied span in the clipboard and insert it in the current position.

To name a few, here are some scenarios where you can use the Copy Tool and the Paste Tool:
1. **Option Selection**: When you need to copy the content of an option in multi-choice questions, you can use the Copy Tool and then the Paste Tool to insert the option content into your response.
2. **Law Article Statement**: When you need to quote a law article, you can use the Copy Tool and then the Paste Tool to insert the law article into your response. 
3. **Terminology Reiteration**: When you need to reiterate a terminology, you can use the Copy Tool and then the Paste Tool to insert the terminology into your response.
4. **Note-Taking**: When you find it necessary to extract some keywords, key phrases, or key sentences from the provided context, you can use the Copy Tool and then the Paste Tool to put down the notes.
5. **URL**: When you find a long URL in the context, it's better to copy it using the Copy Tool and then the Paste Tool to insert it in your response.
6. **Quotation**: When you need to quote a sentence enclosed in either single or double quotes, you can use the Copy Tool and then the Paste Tool to insert the quoted sentence into your response.
7. **Policy Statement**: When you need to quote a policy statement, you can use the Copy Tool and then the Paste Tool to insert the policy statement into your response.
8. **General Text Spans**: Of course, you can use the Copy Tool and the Paste Tool to repeat any text spans that you find necessary to repeat in your response.

Here are the APIs for the Copy Tool and the Paste Tool:
1. **Copy Tool**: `<COPY>[tag="tag1"][description="The description of this copied span."][start=12][end=15]</COPY>.`
    This tool is a function with four parameters: `tag`, `description`, `start`, and `end`.
    In this API, you need to specify the tag, the description of the copied span, and the start and end word positions (including punctuations and the end) of the span that you want to copy.
    You must wrap all the parameters with `<COPY>` and `</COPY>`.
2. **Paste Tool**: `<PASTE>[tag="tag1"]</PASTE>.`
    This tool is a function with one parameter: `tag`, which is strictly the same as the `tag` designed in the previous calling of the Copy Tool.
    Note that there may be multiple calls of the Copy Tool in the previous text, you must use the accurate `tag` to paste the corresponding copied span.
\end{lstlisting}
\end{tcolorbox}

\newpage
\phantomsection
\label{appx:prompt-positionid-cp}
\begin{tcolorbox}[colback=white!95!gray,colframe=gray!50!black,rounded corners,label={prompt-dot}, title={Prompt Template (Few-Shot CP Prompting) (Pre-Generation and Paste Tool Calling) (Continued)}]
\begin{lstlisting}[breaklines=true, xleftmargin=0pt, breakindent=0pt, columns=fullflexible, mathescape]
Here are some examples to illustrate the usage of the Copy Tool and the Paste Tool:

---
**Example 1:**

User: 
What is the capital of France?
a) Berlin  
b) Madrid  
c) Paris  
d) Rome  

Assistant: 
The capital of France is Paris <COPY>[tag="option_c"][description="Option C: Paris"][start=14][end=16]</COPY>. Therefore, the correct answer is <PASTE>[tag="option_c"]</PASTE>.
---
{more demonstrations}

Very Important: 
1. Don't copy the same span over and over again.
2. You must use the Copy Tool and Paste Tool in your response.
3. All the copied spans must be pasted.
4. There may be multiple different Copy Tool Calls, you must accurately paste the right one.
```
\end{lstlisting}
\end{tcolorbox}

\phantomsection
\label{appx:prompt-positionid-cp}
\begin{tcolorbox}[colback=white!95!gray,colframe=gray!50!black,rounded corners,label={prompt-dot}, title={Prompt Template (Few-Shot CP Prompting) (Copy Tool Calling)}]
\begin{lstlisting}[breaklines=true, xleftmargin=0pt, breakindent=0pt, columns=fullflexible, mathescape]
System prompt, during copy tool calling phase:
```
You are a helpful AI assistant.

You are equipped with one tool:
1. **Copy Tool**: If you need to repeat some spans in your response, you have the ability to invoke the Copy Tool. This tool can copy the spans that you want to repeat in your response into the clipboard.

To name a few, here are some scenarios where you can use the Copy Tool and the Paste Tool:
1. **Option Selection**: When you need to copy the content of an option in multi-choice questions, you can use the Copy Tool and then the Paste Tool to insert the option content into your response.
2. **Law Article Statement**: When you need to quote a law article, you can use the Copy Tool and then the Paste Tool to insert the law article into your response. 
3. **Terminology Reiteration**: When you need to reiterate a terminology, you can use the Copy Tool and then the Paste Tool to insert the terminology into your response.
4. **Note-Taking**: When you find it necessary to extract some keywords, key phrases, or key sentences from the provided context, you can use the Copy Tool and then the Paste Tool to put down the notes.
5. **URL**: When you find a long URL in the context, it's better to copy it using the Copy Tool and then the Paste Tool to insert it in your response.
6. **Quotation**: When you need to quote a sentence enclosed in either single or double quotes, you can use the Copy Tool and then the Paste Tool to insert the quoted sentence into your response.
7. **Policy Statement**: When you need to quote a policy statement, you can use the Copy Tool and then the Paste Tool to insert the policy statement into your response.
8. **General Text Spans**: Of course, you can use the Copy Tool and the Paste Tool to repeat any text spans that you find necessary to repeat in your response.

Here are the APIs for the Copy Tool:
1. **Copy Tool**: `<COPY>[tag="tag1"][description="The description of this copied span."][start=12][end=15]</COPY>.`
    This tool is a function with four parameters: `tag`, `description`, `start`, and `end`.
    In this API, you need to specify the tag, the description of the copied span, and the start and end word positions (including punctuations and the end) of the span that you want to copy.
    You must wrap all the parameters with `<COPY>` and `</COPY>`.
\end{lstlisting}
\end{tcolorbox}

\phantomsection
\label{appx:prompt-positionid-cp}
\begin{tcolorbox}[colback=white!95!gray,colframe=gray!50!black,rounded corners,label={prompt-dot}, title={Prompt Template (Few-Shot CP Prompting) (Copy Tool Calling) (Continued)}]
\begin{lstlisting}[breaklines=true, xleftmargin=0pt, breakindent=0pt, columns=fullflexible, mathescape]
Here are some examples to illustrate the usage of the Copy Tool:

---
**Example 1:**

User: 
What is the capital of France?
a) Berlin  
b) Madrid  
c) Paris  
d) Rome  
The capital of France is <COPY>

Assistant: 
[tag="option_c"][description="Option C: Paris"][start=14][end=16]</COPY>
---
{more demonstrations}

Very Important: 
1. Don't copy the same span over and over again.
2. You must use the Copy Tool in your response.
```
\end{lstlisting}
\end{tcolorbox}

\twocolumn

\section{Evaluation Details}
\label{appx:evaluation}

We provide a more detailed explanation of the evaluation metrics mentioned in \S\ref{sec:exp_setup}.

\paragraph{Rouge-L} We use this metric to evaluate the quality of the generated responses. Rouge-L score is calculated through \textit{huggingface evaluate library} \footnote{https://huggingface.co/docs/evaluate/index}.

\paragraph{Mean Absolute Error (MAE)} MAE quantifies the discrepancy between the length of the model's generated response and the instructed length, based on the L1 norm. In our experiments, MAE is formally defined as follows:
    \begin{equation}
        MAE = \frac{1}{N} \sum^{n}_{i=1}{|len(pred_i) - len(label_i)|} \nonumber
    \end{equation}

\paragraph{Perlexity (PPL)} We utilize LLaMA2 7B ~\citep{touvron2023llama} as reference model, and employ the calculation tool of \textit{huggingface evaluate library}.

\paragraph{CP Success Rate} For a given sample, the full response process is considered successful only when: (1) the Copy tool is utilized at least once; (2) Every copied content is used by the Paste tool; (3) Every content called by the Paste tool exists in the Clipboard. Otherwise, the process is considered as a failure. This metric evaluates the efficiency of the copy and paste tool usage.

\paragraph{Clarity, Accuracy, and Consistency} For the results obtained from Few-shot and PositionID CP prompting, we instruct GPT-4o to perform pairwise comparisons with the responses from Zero-shot Non-CP prompting and report the win rates for each method.

\paragraph{Tool-use Proficiency} For responses from Few-shot and PositionID CP prompting, The GPT-4o model rates the tool use from 1 to 5 points, measuring: (i) the usefulness of the copied fragment; (ii) the accuracy of the paste position. During the evaluation, we score the response by comparing the results from Few-shot and PositionID CP prompting.

\onecolumn

\section{LenCtrl-Bench Examples}
\label{appx:lenctrl-examples}

\rule{\textwidth}{0.4pt}
\fbox{LenCtrl-Bench (word-level) (1)}
\begin{lstlisting}[breaklines=true, breakindent=0pt, columns=fullflexible, mathescape]
"instruction": "Rangers say they will \"vigorously pursue\" former chief executive Charles Green for legal costs. Present your thoughts in 230 words.",
"constraint": "230 words",
"level": "word",
"data_source": "wikihow",
"response": "Green claimed his contract entitled him to legal expenses for any court action arising from his time at Ibrox.\nBut the Championship leaders have announced that an appeal from Green's legal team against an earlier court ruling on his expenses has failed.\nHe will now have to pay for his own defence against charges relating to the acquisition of Rangers in 2012.\nA Rangers statement read: \"We are pleased to advise that the appeal by Charles Green against the decision of the Lord Ordinary to dismiss his application to have Rangers International Football Club PLC meet the expenses of the criminal proceedings brought against him has been refused.\n\"Lady Dorrian, Lord Bracadale and Lord Malcolm sitting in the Inner House of the Court of Session unanimously reached this decision.\"\nRangers chairman Dave King added: \"We are unsurprised but delighted that this latest court success brings to an end the unjustifiable claims by Charles Green.\n\"He is now totally responsible for the cost of defending himself in the criminal case brought against him for his dealings with the club.\n\"Furthermore, we will vigorously pursue Charles Green for recovery of the legal costs that we incurred including the £50,000 lodged with the court as a caution.\"\nGreen, who left his Rangers role in April 2013, had his initial legal bid to force the club to pay his legal fees rejected in November.",
"response_ids": "Green 1 claimed 2 his 3 contract 4 entitled 5 him 6 to 7 legal 8 expenses 9 for 10 any 11 court 12 action 13 arising 14 from 15 his 16 time 17 at 18 Ibrox 19 . \n But 20 the 21 Championship 22 leaders 23 have 24 announced 25 that 26 an 27 appeal 28 from 29 Green's 30 legal 31 team 32 against 33 an 34 earlier 35 court 36 ruling 37 on 38 his 39 expenses 40 has 41 failed 42 . \n He 43 will 44 now 45 have 46 to 47 pay 48 for 49 his 50 own 51 defence 52 against 53 charges 54 relating 55 to 56 the 57 acquisition 58 of 59 Rangers 60 in 61 2012 62 . \n A 63 Rangers 64 statement 65 read 66 : \" We 67 are 68 pleased 69 to 70 advise 71 that 72 the 73 appeal 74 by 75 Charles 76 Green 77 against 78 the 79 decision 80 of 81 the 82 Lord 83 Ordinary 84 to 85 dismiss 86 his 87 application 88 to 89 have 90 Rangers 91 International 92 Football 93 Club 94 PLC 95 meet 96 the 97 expenses 98 of 99 the 100 criminal 101 proceedings 102 brought 103 against 104 him 105 has 106 been 107 refused 108 . \n \" Lady 109 Dorrian 110 , Lord 111 Bracadale 112 and 113 Lord 114 Malcolm 115 sitting 116 in 117 the 118 Inner 119 House 120 of 121 the 122 Court 123 of 124 Session 125 unanimously 126 reached 127 this 128 decision 129 . \" \n Rangers 130 chairman 131 Dave 132 King 133 added 134 : \" We 135 are 136 unsurprised 137 but 138 delighted 139 that 140 this 141 latest 142 court 143 success 144 brings 145 to 146 an 147 end 148 the 149 unjustifiable 150 claims 151 by 152 Charles 153 Green 154 . \n \" He 155 is 156 now 157 totally 158 responsible 159 for 160 the 161 cost 162 of 163 defending 164 himself 165 in 166 the 167 criminal 168 case 169 brought 170 against 171 him 172 for 173 his 174 dealings 175 with 176 the 177 club 178 . \n \" Furthermore 179 , we 180 will 181 vigorously 182 pursue 183 Charles 184 Green 185 for 186 recovery 187 of 188 the 189 legal 190 costs 191 that 192 we 193 incurred 194 including 195 the 196 £50,000 197 lodged 198 with 199 the 200 court 201 as 202 a 203 caution 204 . \" \n Green 205 , who 206 left 207 his 208 Rangers 209 role 210 in 211 April 212 2013 213 , had 214 his 215 initial 216 legal 217 bid 218 to 219 force 220 the 221 club 222 to 223 pay 224 his 225 legal 226 fees 227 rejected 228 in 229 November 230 . "
\end{lstlisting}

\rule{\textwidth}{0.4pt}
\fbox{LenCtrl-Bench (word-level) (2)}
\begin{lstlisting}[breaklines=true, breakindent=0pt, columns=fullflexible, mathescape]
"instruction": "How does the internet affect our everyday lives? Offer your insight in just 82 words.",
"constraint": "82 words",
"level": "word",
"data_source": "stanford_alpaca",
"response": "The internet has had a huge impact on our lives, most notably our ability to access information and entertainment, stay connected to family and friends, and work and shop from home. It also provides a forum for sharing opinions and exchanging ideas, which can result in greater awareness, transparency, and accountability. Moreover, it can be a platform for developing educational opportunities, taking part in online communities, and participating in online purchases, which can help to create more jobs and facilitate economic growth.",
"response_ids": "The 1 internet 2 has 3 had 4 a 5 huge 6 impact 7 on 8 our 9 lives 10 , most 11 notably 12 our 13 ability 14 to 15 access 16 information 17 and 18 entertainment 19 , stay 20 connected 21 to 22 family 23 and 24 friends 25 , and 26 work 27 and 28 shop 29 from 30 home 31 . It 32 also 33 provides 34 a 35 forum 36 for 37 sharing 38 opinions 39 and 40 exchanging 41 ideas 42 , which 43 can 44 result 45 in 46 greater 47 awareness 48 , transparency 49 , and 50 accountability 51 . Moreover 52 , it 53 can 54 be 55 a 56 platform 57 for 58 developing 59 educational 60 opportunities 61 , taking 62 part 63 in 64 online 65 communities 66 , and 67 participating 68 in 69 online 70 purchases 71 , which 72 can 73 help 74 to 75 create 76 more 77 jobs 78 and 79 facilitate 80 economic 81 growth 82 . "
\end{lstlisting}

\rule{\textwidth}{0.4pt}
\fbox{LenCtrl-Bench (word-level) (3)}
\begin{lstlisting}[breaklines=true, breakindent=0pt, columns=fullflexible, mathescape]
"instruction": "What type of response does a server provide when it receives an HTTP request? Elaborate in 50 words only.",
"constraint": "50 words",
"level": "word",
"data_source": "stanford_alpaca",
"response": "When a server receives an HTTP request, it responds with an HTTP response, which includes a status code and headers. The status code is a three-digit number that indicates the result of the request, while the headers provide additional information such as the content type and length of the response.",
"response_ids": "When 1 a 2 server 3 receives 4 an 5 HTTP 6 request 7 , it 8 responds 9 with 10 an 11 HTTP 12 response 13 , which 14 includes 15 a 16 status 17 code 18 and 19 headers 20 . The 21 status 22 code 23 is 24 a 25 three-digit 26 number 27 that 28 indicates 29 the 30 result 31 of 32 the 33 request 34 , while 35 the 36 headers 37 provide 38 additional 39 information 40 such 41 as 42 the 43 content 44 type 45 and 46 length 47 of 48 the 49 response 50 . "
\end{lstlisting}

\rule{\textwidth}{0.4pt}
\fbox{LenCtrl-Bench (sentence-level) (1)}
\begin{lstlisting}[breaklines=true, breakindent=0pt, columns=fullflexible, mathescape]
"instruction": "Is the milk produced by a hippopotamus pink in color? With a response length of 3 sentences.",
"constraint": "3 sentences",
"level": "sentence",
"data_source": "openhermes2.5/misconception",
"response": "No, the milk produced by a hippopotamus is not pink. It is typically white or beige in color. The misconception arises due to the hipposudoric acid, a red pigment found in hippo skin secretions, which people mistakenly assume affects the color of their milk.",
"response_ids": "No, the milk produced by a hippopotamus is not pink. <1> It is typically white or beige in color. <2> The misconception arises due to the hipposudoric acid, a red pigment found in hippo skin secretions, which people mistakenly assume affects the color of their milk. <3>"
\end{lstlisting}

\rule{\textwidth}{0.4pt}
\fbox{LenCtrl-Bench (sentence-level) (2)}
\begin{lstlisting}[breaklines=true, breakindent=0pt, columns=fullflexible, mathescape]
"instruction": "Give an example of a bias that could exist in an AI algorithm. Comprising 3 sentences in the response.",
"constraint": "3 sentences",
"level": "sentence",
"data_source": "stanford_alpaca",
"response": "An example of a bias that could exist in an AI algorithm is gender bias. This can be due to gender data used to train the algorithm or the underlying assumptions of the algorithm, which can lead to gender-either outcomes, such as sorting resumes according to gender or assigning job tasks and deadlines according to gender. This type of bias is largely unintentional and may be hard to identify, but can result in systemic gender discrimination.",
"response_ids": "An example of a bias that could exist in an AI algorithm is gender bias. <1> This can be due to gender data used to train the algorithm or the underlying assumptions of the algorithm, which can lead to gender-either outcomes, such as sorting resumes according to gender or assigning job tasks and deadlines according to gender. <2> This type of bias is largely unintentional and may be hard to identify, but can result in systemic gender discrimination. <3>"
\end{lstlisting}

\rule{\textwidth}{0.4pt}
\fbox{LenCtrl-Bench (paragraph-level)}
\begin{lstlisting}[breaklines=true, breakindent=0pt, columns=fullflexible, mathescape]
"instruction": "Write a long and very detailed tutorial on \"Decrease the Chance of Birth Defects\", in the style of WikiHow. Include in depth explanations for each step and how it helps achieve the desired outcome, inluding key tips and guidelines. Ensure clarity and practicality, allowing readers to easily follow and apply the instructions. Do not use images. Enrich your response with 12 paragraphs.",
"constraint": "12 paragraphs",
"level": "paragraph",
"data_source": "wikihow",
"response": " Title: Decrease the Chance of Birth Defects\n\nIntroduction:\nBirth defects can be a source of great concern for expectant parents. While not all birth defects can be prevented, there are several steps you can take to reduce the risk. This comprehensive guide will provide in-depth information on how to decrease the chance of birth defects, with clear and practical instructions.\n\nStep 1: Prepare Your Body Before Pregnancy\nKey Tip: A healthy body is more likely to support a healthy pregnancy. Start preparing your body months before attempting to conceive by adopting these habits:\n\n* Maintain a well-balanced diet rich in fruits, vegetables, lean proteins, whole grains, and low-fat dairy products. These foods contain essential nutrients like folic acid, calcium, iron, and vitamin D that promote fetal growth and development.\n* Take prenatal vitamins containing at least 400 mcg of folic acid daily. Folic acid reduces the risk of neural tube defects (NTDs) when taken prior to conception and during early pregnancy.\n* Reach and maintain a healthy weight. Obesity increases the likelihood of gestational diabetes, high blood pressure, and other complications that may contribute to birth defects. Consult a healthcare professional about appropriate weight goals.\n\nStep 2: Manage Existing Medical Conditions\nGuideline: If you have chronic medical conditions such as diabetes or epilepsy, work closely with your doctor to manage them effectively before becoming pregnant. Properly controlling these conditions decreases the risk of related birth defects.\n\nStep 3: Avoid Harmful Substances\nKey Tips: Refrain from consuming alcohol, tobacco, recreational drugs, and certain medications known to increase the risk of birth defects. Speak with your physician about any prescription medication usage and potential alternatives.\n\nStep 4: Get Vaccinated\nImportance: Certain vaccines protect against infections that could harm an unborn child if contracted during pregnancy. Discuss recommended immunizations with your healthcare provider before trying to conceive.\n\nStep 5: Limit Environmental Risks\nPrecautions: Minimize exposure to harmful environmental factors, including lead paint, radiation, pesticides, solvents, mercury, arsenic, and infectious agents present in soil, water, or air. Use protective gear when necessary and avoid contact with contaminated materials.\n\nStep 6: Practice Food Safety\nRecommendation: Adhere to safe food handling practices to prevent infection from listeria, salmonella, and toxoplasma gondii-pathogens associated with miscarriage, stillbirth, and severe birth defects. Follow these precautions:\n\n* Cook meats thoroughly; avoid deli meat unless heated until steaming hot.\n* Clean raw produce meticulously.\n* Keep uncooked meats separate from ready-to-eat foods.\n* Store leftovers promptly in the refrigerator.\n* Avoid soft cheeses made from unpasteurized milk.\n\nStep 7: Regular Checkups During Pregnancy\nMedical Attention: Schedule regular prenatal appointments to monitor both maternal health and fetal development throughout pregnancy. Early detection and management of issues minimize risks to the baby.\n\nConclusion:\nWhile no method guarantees prevention of birth defects, following this extensive guideline significantly decreases the chances of encountering problems during pregnancy. By adhering to these recommendations, prospective mothers create optimal conditions for carrying their babies to term without complications.",
"response_ids": " Title: Decrease the Chance of Birth Defects <1>\nIntroduction:\nBirth defects can be a source of great concern for expectant parents. While not all birth defects can be prevented, there are several steps you can take to reduce the risk. This comprehensive guide will provide in. depth information on how to decrease the chance of birth defects, with clear and practical instructions. <2>\nStep 1: Prepare Your Body Before Pregnancy\nKey Tip: A healthy body is more likely to support a healthy pregnancy. Start preparing your body months before attempting to conceive by adopting these habits: <3>\n* Maintain a well. balanced diet rich in fruits, vegetables, lean proteins, whole grains, and low. fat dairy products. These foods contain essential nutrients like folic acid, calcium, iron, and vitamin D that promote fetal growth and development.\n* Take prenatal vitamins containing at least 400 mcg of folic acid daily. Folic acid reduces the risk of neural tube defects (NTDs) when taken prior to conception and during early pregnancy.\n* Reach and maintain a healthy weight. Obesity increases the likelihood of gestational diabetes, high blood pressure, and other complications that may contribute to birth defects. Consult a healthcare professional about appropriate weight goals. <4>\nStep 2: Manage Existing Medical Conditions\nGuideline: If you have chronic medical conditions such as diabetes or epilepsy, work closely with your doctor to manage them effectively before becoming pregnant. Properly controlling these conditions decreases the risk of related birth defects. <5>\nStep 3: Avoid Harmful Substances\nKey Tips: Refrain from consuming alcohol, tobacco, recreational drugs, and certain medications known to increase the risk of birth defects. Speak with your physician about any prescription medication usage and potential alternatives. <6>\nStep 4: Get Vaccinated\nImportance: Certain vaccines protect against infections that could harm an unborn child if contracted during pregnancy. Discuss recommended immunizations with your healthcare provider before trying to conceive. <7>\nStep 5: Limit Environmental Risks\nPrecautions: Minimize exposure to harmful environmental factors, including lead paint, radiation, pesticides, solvents, mercury, arsenic, and infectious agents present in soil, water, or air. Use protective gear when necessary and avoid contact with contaminated materials. <8>\nStep 6: Practice Food Safety\nRecommendation: Adhere to safe food handling practices to prevent infection from listeria, salmonella, and toxoplasma gondii-pathogens associated with miscarriage, stillbirth, and severe birth defects. Follow these precautions: <9>\n* Cook meats thoroughly; avoid deli meat unless heated until steaming hot.\n* Clean raw produce meticulously.\n* Keep uncooked meats separate from ready. to. eat foods.\n* Store leftovers promptly in the refrigerator.\n* Avoid soft cheeses made from unpasteurized milk. <10>\nStep 7: Regular Checkups During Pregnancy\nMedical Attention: Schedule regular prenatal appointments to monitor both maternal health and fetal development throughout pregnancy. Early detection and management of issues minimize risks to the baby. <11>\nConclusion:\nWhile no method guarantees prevention of birth defects, following this extensive guideline significantly decreases the chances of encountering problems during pregnancy. By adhering to these recommendations, prospective mothers create optimal conditions for carrying their babies to term without complications. <12>\n"
\end{lstlisting}

\section{CP-Bench Examples} 
\label{appx:cpbench-examples}

\rule{\textwidth}{0.4pt}
\fbox{CP-Bench (Terminology Reiteration)}
\begin{lstlisting}[breaklines=true, breakindent=0pt, columns=fullflexible, mathescape]
{"instruction": "Can you explain how gene expression is regulated in cells?", "instruction_ids": "Can[1] you[2] explain[3] how[4] gene[5] expression[6] is[7] regulated[8] in[9] cells[10] ?[11] ", "type": "single-copy", "cp_source": "input", "data_source": "", "principle": "terminology reiteration", "cp_density": "2.78", "response": "Certainly! Gene expression is regulated through mechanisms such as transcriptional control, where transcription factors bind to specific DNA sequences to activate or repress the transcription of genes, and post-transcriptional modifications, which include processes like mRNA splicing and polyadenylation to produce mature mRNA. This ensures that <COPY>[tag=Biological process][description=Regulation of gene activity][start=5][end=6]</COPY><PASTE>[tag=Biological process]</PASTE> is finely tuned according to the cell's needs.", "response_ids": "Certainly[12] ![13] Gene[14] expression[15] is[16] regulated[17] through[18] mechanisms[19] such[20] as[21] transcriptional[22] control[23] ,[24] where[25] transcription[26] factors[27] bind[28] to[29] specific[30] DNA[31] sequences[32] to[33] activate[34] or[35] repress[36] the[37] transcription[38] of[39] genes[40] ,[41] and[42] post-transcriptional[43] modifications[44] ,[45] which[46] include[47] processes[48] like[49] mRNA[50] splicing[51] and[52] polyadenylation[53] to[54] produce[55] mature[56] mRNA[57] .[58] This[59] ensures[60] that[61] <COPY>[tag=Biological process][description=Regulation of gene activity][start=5][end=6]</COPY><PASTE>[tag=Biological process]</PASTE> is[64] finely[65] tuned[66] according[67] to[68] the[69] cell's[70] needs[71] .[72] ", "response_pure_text": "Certainly! Gene expression is regulated through mechanisms such as transcriptional control, where transcription factors bind to specific DNA sequences to activate or repress the transcription of genes, and post-transcriptional modifications, which include processes like mRNA splicing and polyadenylation to produce mature mRNA. This ensures that gene expression is finely tuned according to the cell's needs."}
\end{lstlisting}

\rule{\textwidth}{0.4pt}
\fbox{CP-Bench (Option Selection)}
\begin{lstlisting}[breaklines=true, breakindent=0pt, columns=fullflexible, mathescape]
"instruction": "In the context of world history, what was the primary cause of the fall of the Western Roman Empire in the 5th century AD? A. The rise of Christianity B. Economic instability and inflation C. The sacking of Rome by the Visigoths D. The invasion and migration of the Germanic tribes E. All of the above ", "instruction_ids": "In[1] the[2] context[3] of[4] world[5] history[6] ,[7] what[8] was[9] the[10] primary[11] cause[12] of[13] the[14] fall[15] of[16] the[17] Western[18] Roman[19] Empire[20] in[21] the[22] 5th[23] century[24] AD[25] ?[26] A[27] .[28] The[29] rise[30] of[31] Christianity[32] B[33] .[34] Economic[35] instability[36] and[37] inflation[38] C.[39] The[40] sacking[41] of[42] Rome[43] by[44] the[45] Visigoths[46] D.[47] The[48] invasion[49] and[50] migration[51] of[52] the[53] Germanic[54] tribes[55] E.[56] All[57] of[58] the[59] above[60] ", "type": "single-copy", "cp_source": "input", "data_source": "openhermes2.5/multiple_choice", "principle": "option selection", "cp_density": "13.04", "response": "<COPY>[tag=Answer option D][description=The chosen answer for Rome's fall][start=47][end=55]</COPY><PASTE>[tag=Answer option D]</PASTE> ", "response_ids": "<COPY>[tag=Answer option D][description=The chosen answer for Rome's fall][start=47][end=55]</COPY><PASTE>[tag=Answer option D]</PASTE> ", "response_pure_text": "D. The invasion and migration of the Germanic tribes "
\end{lstlisting}

\rule{\textwidth}{0.4pt}
\fbox{CP-Bench (Law Article Statement)}
\begin{lstlisting}[breaklines=true, breakindent=0pt, columns=fullflexible, mathescape]
"instruction": "8. The applicant is the owner of a flat in Milan, which she had let to L.D.Z. 9. In a registered letter of 6 June 1984, the applicant informed the tenant that she intended to terminate the lease on expiry of the term on 29 December 1984 and asked her to vacate the premises by that date. 10. On 11 February 1985, she served a notice to quit on the tenant, but she refused to leave. 11. In a writ served on the tenant on 19 February 1985, the applicant reiterated her intention to terminate the lease and summoned the tenant to appear before the Milan Magistrate. 12. By a decision of 27 February 1985, which was made enforceable on 14 March 1985, the Milan Magistrate upheld the validity of the notice to quit and ordered that the premises be vacated by 27 February 1986. 13. On 23 January 1986, the applicant served notice on the tenant requiring her to vacate the premises. 14. On 7 March 1986, she served notice on the tenant informing her that the order for possession would be enforced by a bailiff on 18 April 1986. 15. Between 18 April 1986 and 18 June 1992 the bailiff made 23 attempts to recover possession. Each attempt proved unsuccessful, as, under the statutory provisions providing for the suspension or the staggering of evictions, the applicant was not entitled to police assistance in enforcing the order for possession. 16. Thereafter, the applicant decided not to pursue the enforcement proceedings, in order to avoid useless costs, given the lack of prospects of obtaining the assistance of the police. 17. On 13 April 1996 the applicant repossessed the flat, which the tenant vacated in pursuance of an agreement reached with the applicant.  According to the above cases, which ECHR articles were violated. Please select the correct answers from the following options:  \"Article 3: Prohibition of torture\", \"Article 6: Right to a fair trial\", \"Article 10: Freedom of expression\", \"Article 14: Prohibition of discrimination\", \"Article 1 of Protocol 1: Protection of property\", ", "instruction_ids": "8[1] .[2] The[3] applicant[4] is[5] the[6] owner[7] of[8] a[9] flat[10] in[11] Milan[12] ,[13] which[14] she[15] had[16] let[17] to[18] L.D.Z[19] .[20] 9[21] .[22] In[23] a[24] registered[25] letter[26] of[27] 6[28] June[29] 1984[30] ,[31] the[32] applicant[33] informed[34] the[35] tenant[36] that[37] she[38] intended[39] to[40] terminate[41] the[42] lease[43] on[44] expiry[45] of[46] the[47] term[48] on[49] 29[50] December[51] 1984[52] and[53] asked[54] her[55] to[56] vacate[57] the[58] premises[59] by[60] that[61] date[62] .[63] 10[64] .[65] On[66] 11[67] February[68] 1985[69] ,[70] she[71] served[72] a[73] notice[74] to[75] quit[76] on[77] the[78] tenant[79] ,[80] but[81] she[82] refused[83] to[84] leave[85] .[86] 11[87] .[88] In[89] a[90] writ[91] served[92] on[93] the[94] tenant[95] on[96] 19[97] February[98] 1985[99] ,[100] the[101] applicant[102] reiterated[103] her[104] intention[105] to[106] terminate[107] the[108] lease[109] and[110] summoned[111] the[112] 
tenant[113] to[114] appear[115] before[116] the[117] Milan[118] Magistrate[119] .[120] 12[121] .[122] By[123] a[124] decision[125] of[126] 27[127] February[128] 1985[129] ,[130] which[131] was[132] made[133] enforceable[134] on[135] 14[136] March[137] 1985[138] ,[139] the[140] Milan[141] Magistrate[142] upheld[143] the[144] validity[145] of[146] the[147] notice[148] to[149] quit[150] and[151] ordered[152] that[153] the[154] premises[155] be[156] vacated[157] by[158] 27[159] February[160] 1986[161] .[162] 13[163] .[164] On[165] 23[166] January[167] 1986[168] ,[169] the[170] applicant[171] served[172] notice[173] on[174] the[175] tenant[176] requiring[177] her[178] to[179] vacate[180] the[181] premises[182] .[183] 14[184] .[185] On[186] 7[187] March[188] 1986[189] ,[190] she[191] served[192] notice[193] on[194] the[195] tenant[196] informing[197] her[198] that[199] the[200] order[201] for[202] possession[203] would[204] be[205] enforced[206] by[207] a[208] bailiff[209] on[210] 18[211] April[212] 1986[213] .[214] 15[215] .[216] Between[217] 18[218] April[219] 1986[220] and[221] 18[222] June[223] 1992[224] the[225] bailiff[226] made[227] 23[228] attempts[229] to[230] recover[231] possession[232] .[233] Each[234] attempt[235] proved[236] unsuccessful[237] ,[238] as[239] ,[240] under[241] the[242] statutory[243] provisions[244] providing[245] for[246] the[247] suspension[248] or[249] the[250] staggering[251] of[252] evictions[253] ,[254] the[255] applicant[256] was[257] not[258] entitled[259] to[260] police[261] assistance[262] 
in[263] enforcing[264] the[265] order[266] for[267] possession[268] .[269] 16[270] .[271] Thereafter[272] ,[273] the[274] applicant[275] decided[276] not[277] to[278] pursue[279] the[280] enforcement[281] proceedings[282] ,[283] in[284] order[285] to[286] avoid[287] useless[288] costs[289] ,[290] given[291] the[292] lack[293] of[294] prospects[295] of[296] obtaining[297] the[298] assistance[299] of[300] the[301] police[302] .[303] 17[304] .[305] On[306] 13[307] April[308] 1996[309] the[310] applicant[311] repossessed[312] the[313] flat[314] ,[315] which[316] the[317] tenant[318] vacated[319] in[320] pursuance[321] of[322] an[323] agreement[324] reached[325] with[326] the[327] applicant[328] .[329] According[330] to[331] the[332] above[333] cases[334] ,[335] which[336] ECHR[337] articles[338] were[339] violated[340] .[341] Please[342] select[343] the[344] correct[345] answers[346] from[347] the[348] following[349] options[350] :[351] \"[352] Article[353] 3[354] :[355] Prohibition[356] of[357] torture[358] \"[359] ,[360] \"[361] Article[362] 6[363] :[364] Right[365] to[366] a[367] fair[368] trial[369] \"[370] ,[371] \"[372] Article[373] 10[374] :[375] Freedom[376] of[377] expression[378] \"[379] ,[380] \"[381] Article[382] 14[383] :[384] Prohibition[385] of[386] discrimination[387] \"[388] ,[389] \"[390] Article[391] 1[392] of[393] Protocol[394] 1[395] :[396] Protection[397] of[398] property[399] \"[400] ,[401] ", "type": "multi-copy", "cp_source": "input", "data_source": "LexGLUE", "principle": "law article statement", "cp_density": "3.97", "response": "The answer is: \"<COPY>[tag=ECHR Articles Violated][description=Right to a fair trial][start=362][end=369]</COPY><PASTE>[tag=ECHR Articles Violated]</PASTE>\", \"<COPY>[tag=ECHR article][description=Refers to property protection violation][start=391][end=399]</COPY><PASTE>[tag=ECHR article]</PASTE>\". ", "response_ids": "The[402] answer[403] is[404] :[405] \"[406] <COPY>[tag=ECHR Articles Violated][description=Right to a fair trial][start=362][end=369]</COPY><PASTE>[tag=ECHR Articles Violated]</PASTE> \"[415] ,[416] \"[417] <COPY>[tag=ECHR article][description=Refers to property protection violation][start=391][end=399]</COPY><PASTE>[tag=ECHR article]</PASTE> \"[427] .[428] ", "response_pure_text": "The answer is: \"Article 6: Right to a fair trial\", \"Article 1 of Protocol 1: Protection of property\". "
\end{lstlisting}

\rule{\textwidth}{0.4pt}
\fbox{CP-Bench (Policy Statement)}
\begin{lstlisting}[breaklines=true, breakindent=0pt, columns=fullflexible, mathescape]
"instruction": "Online Behavioral Advertising We or our advertising service providers may use certain information about your activities on our Sites, such as pages visited and search key words entered to help determine which of our advertisements or offers may be of interest to you. We limit access and collection of information for specific purposes by advertising service providers. We may use this online information for online and offline advertising. Does the website allow accessing the user's data? If yes, why? ", "instruction_ids": "Online[1] Behavioral[2] Advertising[3] We[4] or[5] our[6] advertising[7] service[8] providers[9] may[10] use[11] certain[12] information[13] about[14] your[15] activities[16] on[17] our[18] Sites[19] ,[20] such[21] as[22] pages[23] visited[24] and[25] search[26] key[27] words[28] entered[29] to[30] help[31] determine[32] which[33] of[34] our[35] advertisements[36] or[37] offers[38] may[39] be[40] of[41] interest[42] to[43] you[44] .[45] We[46] limit[47] access[48] and[49] collection[50] of[51] information[52] for[53] specific[54] purposes[55] by[56] advertising[57] service[58] providers[59] .[60] We[61] may[62] use[63] this[64] online[65] information[66] for[67] online[68] and[69] offline[70] advertising[71] .[72] Does[73] the[74] website[75] allow[76] accessing[77] the[78] user's[79] data[80] ?[81] If[82] yes[83] ,[84] why[85] ?[86] ", "type": "single-copy", "cp_source": "input", "data_source": "PolicyQA", "principle": "policy statement", "cp_density": "14.85", "response": "<COPY>[tag=Limited data access][description=Controlled collection for advertising purposes][start=46][end=60]</COPY><PASTE>[tag=Limited data access]</PASTE> ", "response_ids": "<COPY>[tag=Limited data access][description=Controlled collection for advertising purposes][start=46][end=60]</COPY><PASTE>[tag=Limited data access]</PASTE> ", "response_pure_text": "We limit access and collection of information for specific purposes by advertising service providers. "
\end{lstlisting}

\rule{\textwidth}{0.4pt}
\fbox{CP-Bench (Note Taking)}
\begin{lstlisting}[breaklines=true, breakindent=0pt, columns=fullflexible, mathescape]
"instruction": "\"The analysis of healthcare markets and the economic behavior within these markets is crucial for understanding how policies can be tailored to improve healthcare outcomes. Health economics not only evaluates the efficiency and effectiveness of healthcare systems but also examines how resources are allocated, the impact of health insurance, and the socioeconomic factors that affect health. Through econometric analyses and economic modeling, researchers can identify cost-effective strategies that enhance access to healthcare services while controlling expenditure growth. For instance, studies often focus on the relationship between insurance coverage and access to care, exploring how different insurance designs can mitigate risk and improve policyholder wellbeing.\" Please excerpt what you think is important from the original text.", "instruction_ids": "\"[1] The[2] analysis[3] of[4] healthcare[5] markets[6] and[7] the[8] economic[9] behavior[10] within[11] these[12] markets[13] is[14] crucial[15] for[16] understanding[17] how[18] policies[19] can[20] be[21] tailored[22] to[23] improve[24] healthcare[25] outcomes[26] .[27] Health[28] economics[29] not[30] only[31] evaluates[32] the[33] efficiency[34] and[35] effectiveness[36] of[37] healthcare[38] systems[39] but[40] also[41] examines[42] how[43] resources[44] are[45] allocated[46] ,[47] the[48] impact[49] of[50] health[51] insurance[52] ,[53] and[54] the[55] socioeconomic[56] factors[57] that[58] affect[59] health[60] .[61] Through[62] econometric[63] analyses[64] and[65] economic[66] modeling[67] ,[68] researchers[69] can[70] identify[71] cost-effective[72] strategies[73] that[74] enhance[75] access[76] to[77] healthcare[78] services[79] while[80] controlling[81] expenditure[82] growth[83] .[84] For[85] instance[86] ,[87] studies[88] often[89] focus[90] on[91] the[92] relationship[93] between[94] insurance[95] coverage[96] and[97] access[98] to[99] care[100] ,[101] exploring[102] how[103] different[104] insurance[105] designs[106] can[107] mitigate[108] risk[109] and[110] improve[111] policyholder[112] wellbeing[113] .[114] \"[115] Please[116] excerpt[117] what[118] you[119] think[120] is[121] important[122] from[123] the[124] original[125] text[126] .[127] ", "type": "single-copy", "cp_source": "input", "data_source": "", "principle": "note-taking", "cp_density": "21.12", "response": "<COPY>[tag=Health economics summary][description=Key aspects of health economics][start=28][end=61]</COPY><PASTE>[tag=Health economics summary]</PASTE>", "response_ids": "<COPY>[tag=Health economics summary][description=Key aspects of health economics][start=28][end=61]</COPY><PASTE>[tag=Health economics summary]</PASTE> ", "response_pure_text": "Health economics not only evaluates the efficiency and effectiveness of healthcare systems but also examines how resources are allocated, the impact of health insurance, and the socioeconomic factors that affect health."
\end{lstlisting}

\rule{\textwidth}{0.4pt}
\fbox{CP-Bench (URL)}
\begin{lstlisting}[breaklines=true, breakindent=0pt, columns=fullflexible, mathescape]
"instruction": "url: https://www.biography.com/scientist/albert-einstein  Albert Einstein, a prominent figure in physics, showed prodigious talent from an early age. He started speaking at just one year old and reading complex texts by the age of three. Although he made significant contributions to science, he never received the Nobel Prize in Physics.   At what ages did Albert Einstein start speaking and reading, and did he win a Nobel Prize? Citation? ", "instruction_ids": "url[1] :[2] https[3] :[4] //www.biography.com/scientist/albert-einstein[5] Albert[6] Einstein[7] ,[8] a[9] prominent[10] figure[11] in[12] physics[13] ,[14] showed[15] prodigious[16] talent[17] from[18] an[19] early[20] age[21] .[22] He[23] started[24] speaking[25] at[26] just[27] one[28] year[29] old[30] and[31] reading[32] complex[33] texts[34] by[35] the[36] age[37] of[38] three[39] .[40] Although[41] he[42] made[43] significant[44] contributions[45] to[46] science[47] ,[48] he[49] never[50] received[51] the[52] Nobel[53] Prize[54] in[55] Physics[56] .[57] At[58] what[59] ages[60] did[61] Albert[62] Einstein[63] start[64] speaking[65] and[66] reading[67] ,[68] and[69] did[70] he[71] win[72] a[73] Nobel[74] Prize[75] ?[76] Citation[77] ?[78] ", "type": "single-copy", "cp_source": "input", "data_source": "openhermes2.5/counterfactual_contextual", "principle": "url repeat", "cp_density": "2.54", "response": "Albert Einstein started speaking at the age of one and began reading complex texts by the age of three. He did not receive the Nobel Prize in Physics. Reference(s): url: <COPY>[tag=Einstein Biography URL][description=URL for Albert Einstein's biography.][start=3][end=5]</COPY><PASTE>[tag=Einstein Biography URL]</PASTE> ", "response_ids": "Albert[79] Einstein[80] started[81] speaking[82] at[83] the[84] age[85] of[86] one[87] and[88] began[89] reading[90] complex[91] texts[92] by[93] the[94] age[95] of[96] three[97] .[98] He[99] did[100] not[101] receive[102] the[103] Nobel[104] Prize[105] in[106] Physics[107] .[108] Reference[109] ([110] s[111] )[112] :[113] url[114] :[115] <COPY>[tag=Einstein Biography URL][description=URL for Albert Einstein's biography.][start=3][end=5]</COPY><PASTE>[tag=Einstein Biography URL]</PASTE> ", "response_pure_text": "Albert Einstein started speaking at the age of one and began reading complex texts by the age of three. He did not receive the Nobel Prize in Physics. Reference(s): url: https://www.biography.com/scientist/albert-einstein "
\end{lstlisting}

\rule{\textwidth}{0.4pt}
\fbox{CP-Bench (Quotation)}
\begin{lstlisting}[breaklines=true, breakindent=0pt, columns=fullflexible, mathescape]
"instruction": "Who is the author of the poem \"The Road Not Taken\"? ", "instruction_ids": "Who[1] is[2] the[3] author[4] of[5] the[6] poem[7] \"[8] The[9] Road[10] Not[11] Taken[12] \"[13] ?[14] ", "type": "single-copy", "cp_source": "input", "data_source": "openhermes2.5/general", "principle": "quotation", "cp_density": "13.79", "response": "The author of the poem \"<COPY>[tag=poem title][description=Title of a poem by Robert Frost][start=9][end=12]</COPY><PASTE>[tag=poem title]</PASTE>\" is Robert Frost. ", "response_ids": "The[15] author[16] of[17] the[18] poem[19] \"[20] <COPY>[tag=poem title][description=Title of a poem by Robert Frost][start=9][end=12]</COPY><PASTE>[tag=poem title]</PASTE> \"[25] is[26] Robert[27] Frost[28] .[29] ", "response_pure_text": "The author of the poem \"The Road Not Taken\" is Robert Frost. "
\end{lstlisting}

\rule{\textwidth}{0.4pt}
\fbox{CP-Bench (General Text Spans)}
\begin{lstlisting}[breaklines=true, breakindent=0pt, columns=fullflexible, mathescape]
"instruction": "Physician: In itself, exercise does not cause heart attacks; rather, a sudden increase in an exercise regimen can be a cause. When people of any physical condition suddenly increase their amount of exercise, they also increase their risk of heart attack. As a result, there will be an increased risk of heart attack among employees of this company due to the new health program. The conclusion drawn by the physician follows logically if which one of the following is assumed? A.Employees will abruptly increase their amount of exercise as a result of the new health program. B.The new health program will force employees of all levels of health to exercise regularly. C.The new health program constitutes a sudden change in the company's policy. D.All employees, no matter what their physical condition, will participate in the new health program. ", "instruction_ids": "Physician[1] :[2] In[3] itself[4] ,[5] exercise[6] does[7] not[8] cause[9] heart[10] attacks[11] ;[12] rather[13] ,[14] a[15] sudden[16] increase[17] in[18] an[19] exercise[20] regimen[21] can[22] be[23] a[24] cause[25] .[26] When[27] people[28] of[29] any[30] physical[31] condition[32] suddenly[33] increase[34] their[35] amount[36] of[37] exercise[38] ,[39] they[40] also[41] increase[42] their[43] risk[44] of[45] heart[46] attack[47] .[48] As[49] a[50] result[51] ,[52] there[53] will[54] be[55] an[56] increased[57] risk[58] of[59] heart[60] attack[61] among[62] employees[63] of[64] this[65] company[66] due[67] to[68] the[69] new[70] health[71] program[72] .[73] The[74] conclusion[75] drawn[76] by[77] the[78] physician[79] follows[80] logically[81] if[82] which[83] one[84] of[85] the[86] following[87] is[88] assumed[89] ?[90] A.Employees[91] will[92] abruptly[93] increase[94] their[95] amount[96] of[97] exercise[98] as[99] a[100] result[101] of[102] the[103] new[104] health[105] program[106] .[107] B.The[108] new[109] health[110] program[111] will[112] force[113] employees[114] of[115] all[116] levels[117] of[118] health[119] to[120] exercise[121] regularly[122] .[123] C.The[124] new[125] health[126] program[127] constitutes[128] a[129] sudden[130] change[131] in[132] the[133] company's[134] policy[135] .[136] D.All[137] employees[138] ,[139] no[140] matter[141] what[142] their[143] physical[144] condition[145] ,[146] will[147] participate[148] in[149] the[150] new[151] health[152] program[153] .[154] ", "type": "single-copy", "cp_source": "input", "data_source": "reclor", "principle": "general text spans", "cp_density": "9.34", "response": "Stemming from the question, the answer is: \"<COPY>[tag=Answer choice \"A\"][description=Identifies correct assumption for physician's conclusion.][start=91][end=107]</COPY><PASTE>[tag=Answer choice \"A\"]</PASTE>\" ", "response_ids": "Stemming[155] from[156] the[157] question[158] ,[159] the[160] answer[161] is[162] :[163] \"[164] <COPY>[tag=Answer choice \"A\"][description=Identifies correct assumption for physician's conclusion.][start=91][end=107]</COPY><PASTE>[tag=Answer choice \"A\"]</PASTE> \"[182] ", "response_pure_text": "Stemming from the question, the answer is: \"A.Employees will abruptly increase their amount of exercise as a result of the new health program.\" "
\end{lstlisting}

\newpage

\twocolumn
\section{More Experimental Analysis\label{appx:more-experiments}}

\begin{table}[ht]
    \centering
    \begin{tabular}{c|c|c|c}
    \toprule
    \textbf{Method} & \textbf{Rouge-L} & \textbf{BLEU} & \textbf{METEOR} \\
    \midrule
    \textbf{Zero-Shot} & 22.51 & 15.98 & 27.78 \\
    \textbf{PositionID} & 23.20 & 16.90 & 28.60 \\
    \bottomrule
    \end{tabular}
    \caption{Performance comparison of different prompting methods on LenCtrl-Bench using Rouge-L, BLEU, and METEOR as text quality metrics.}
    \label{tab:more-quality-metrics}
\end{table}

\paragraph{More Metrics for Response Quality.} 
In Table \ref{tab:more-quality-metrics}, we add BLEU~\citep{bleu} and METEOR~\citep{banerjee-lavie-2005-meteor} metrics to further enrich the assessment of text quality of Length Control Prompting. We find that the conclusions drawn from these two new metrics are consistent with those of the ROUGE-L metric in \S\ref{sec:results-positionid-prompting}, all of which demonstrate that the PositionID Prompting method does not compromise the quality of text generation.

We also conduct human evaluation to further assess the response quality, we recruit three graduate students specializing in NLP for annotation. We randomly select 300 responses from GPT-4 PositionID Prompting for word-level length control. The annotators are asked to compare the quality and accuracy of the model responses against the ground-truths, and provide a ``win/lose'' boolean score for each sample. The evaluation criteria cover aspects like text coherence, answer correctness, completeness, and interestingness. The final results are determined by majority voting. We find that the majority-voted win rate was 138/300 (46.0\%), indicating that PositionID Prompting does not significantly influence the quality of the generated responses.

\begin{figure*}[h]
    \centering
    \includegraphics[width=1\linewidth]{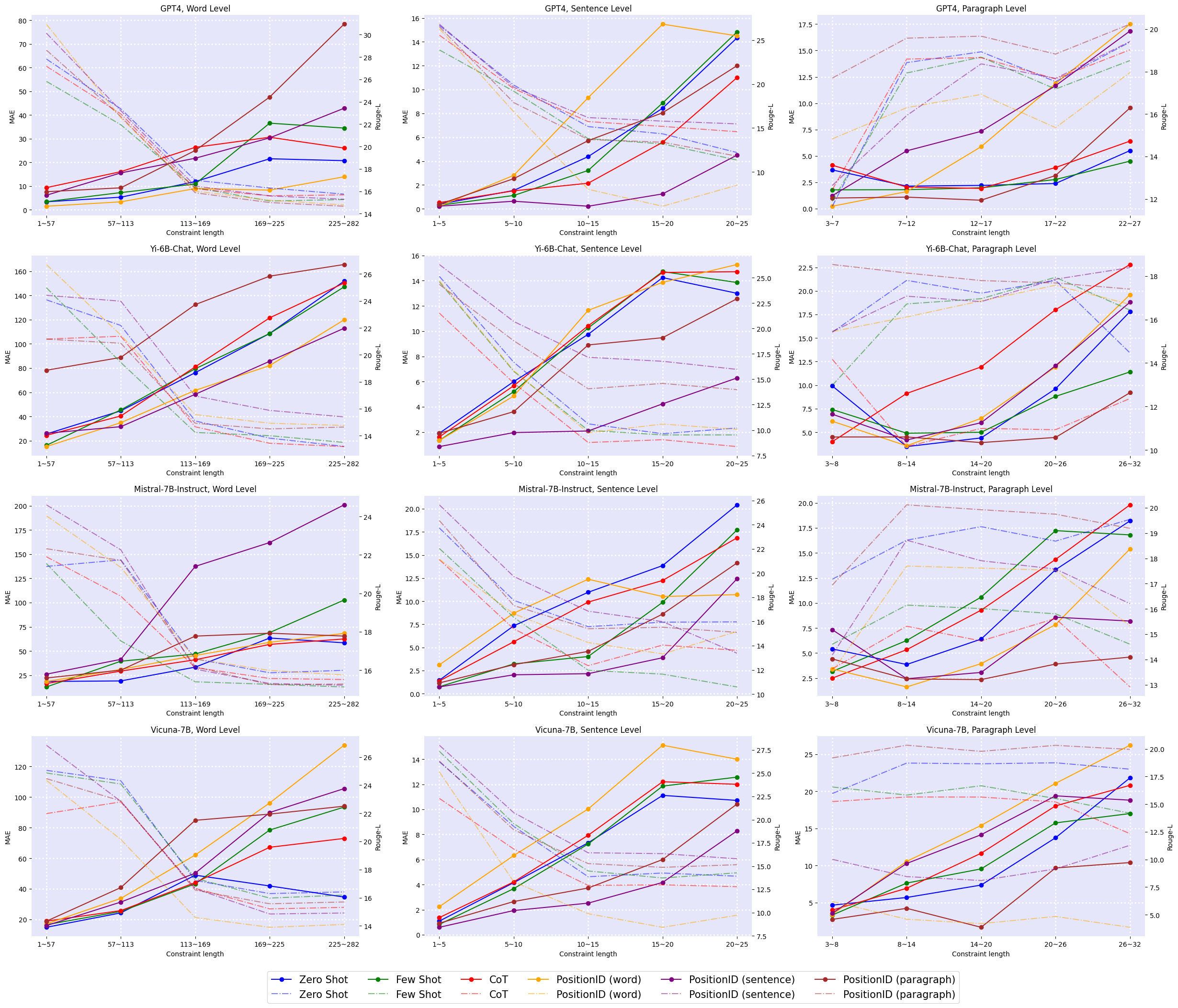}
    \caption{Performance on LenCtrl-Bench with different constraint lengths and different models. The solid line stands for the MAE scores, while the dotted line indicates the Rouge-L scores.}
    \label{fig:total_results}
\end{figure*}

\paragraph{More Results on the Effect of Constraint Length on Prompting.}
\label{appendix:mix_reuslts_constraint} 
In Figure \ref{fig:total_results}, we report more results of different models at different levels.

\begin{table*}[ht]
    \centering
    \begin{tabular}{c|c|c|c|c}
    \toprule
    \textbf{Method} & \textbf{Rouge-L} & \textbf{MAE (word)} & \textbf{MAE (sentence)} & \textbf{MAE (paragraph)} \\
    \midrule
    \textbf{Mistral-7B-Instruct} & 19.85 & 29.64 & 5.23 & 7.19 \\
    \textbf{+CFT} & 21.55 & 29.12 & 5.97 & 7.24 \\
    \textbf{+InstructCTG} & 21.30 & 31.91 & 5.24 & 7.20 \\ \midrule
    \textbf{+PositionID Fine-Tuning} & \textbf{22.13} & \textbf{27.09} & \textbf{4.36} & \textbf{6.88} \\
    \bottomrule
    \end{tabular}
    \caption{The experimental results of fine-tuning methods on LenCtrl-Bench with Mistral-7B-Instruct.}
    \label{tab:mistral-position-ft}
\end{table*}

\paragraph{More Models for PositionID Fine-Tuning.} 
Table \ref{tab:mistral-position-ft} shows the experimental results from fine-tuning Mistral-7B-Instruct-v0.2~\citep{jiang2023mistral}. We can observe that, similar to the experiments conducted with the Yi model in Table \ref{tab:eval-prompts}, PositionID Fine-Tuning is the most effective approach in terms of both text quality and length control.

\begin{table*}[ht]
    \centering
    \begin{tabular}{c|c|c|c|c}
    \toprule
    \textbf{Method} & \textbf{Rouge-L} & \textbf{MAE (word)} & \textbf{MAE (sentence)} & \textbf{MAE (paragraph)} \\
    \midrule
    \textbf{PositionID + Normal} & 19.8 & \textbf{53.8} & \textbf{4.9} & \textbf{6.8} \\
    \textbf{PositionID $\rightarrow$ Normal} & \textbf{20.3} & 58.5 & 5.0 & 7.2 \\
    \bottomrule
    \end{tabular}
    \caption{Comparing mode-mixing training with sequential training.}
    \label{tab:mode-mixing}
\end{table*}

\paragraph{Ablation Experiments on the Mode Mixing for PositionID Fine-Tuning.} In our standard practice, the PositionID fine-tuning is conducted on a mixture of normal-mode data and PositionID-mode data. We introduce an additional experiment that first fine-tunes on PositionID-mode data and then on normal-mode data. We do not consider the normal $\rightarrow$ PositionID training order since the final model must generate responses in normal mode to avoid using position IDs. 
We can see from Table \ref{tab:mode-mixing} that sequential training can cause the model to forget the learned length control knowledge. Therefore, mixed training is a better practice.

\section{FAQs}

\textbf{Question 1: The motivation of the copy-paste task and the necessity of external tool for copy-paste.}

Answer 1: Copy-pasting has always been a crucial aspect of human work, providing at least two key benefits: (1) accelerating writing speed and (2) ensuring the accuracy and consistency of repeated information. Our constructed test set is the first to focus on assessing this capability in LLMs. Our scenario design includes tasks such as copying phone numbers and URLs, which require strict and error-free repetition of existing information. Text generation based on probabilistic methods, as used by GPT, often cannot be trusted completely by users, unlike a deterministic hard-coded copy-paste function.

Furthermore, when the information to be repeated is lengthy, such as in code rewriting tasks, GPT without copy-paste capability tends to take a long time to reproduce the original content, while a simple copy-paste could accomplish this instantly. Our work is the first to emphasize this capability, aiming to inspire subsequent comprehensive and real-world evaluations and optimizations of copy-and-paste functionality.

\textbf{Question 2: PositionID Prompting doubles the input length, which is inefficient.
}

Answer 2: PositionID Prompting is inefficient, which is why we further introduce PositionID Fine-Tuning. The decoding phase for PositionID fine-tuning follows the normal mode without position IDs. The models learn the knowledge of length control from the PositionID mode and generate the response in the normal mode with this acquired knowledge.

Furthermore, for closed-source models, we are often unable or lack the resources to fine-tune the models. This situation presents a trade-off between efficiency and length control. However, in certain cases, users may prioritize length control over efficiency. After all, the model generation processes can be run in the background. Also, users of ChatGPT on web platforms often do not need to worry about the cost of GPT API calling.

The case is similar to CoT Prompting. CoT Prompting is also inefficient compared to direct answering. However, when users care more about the quality and accuracy of the response, the inefficiency can be negligible.

\textbf{Question 3: The intuition on the design of evaluation metrics for copy-paste.
}

Answer 3: From the perspective of copy-paste, a comprehensive evaluation should include: A. Assessing whether the tool was \textbf{successfully} invoked; B. Evaluating whether the tool was \textbf{placed correctly} after being invoked; C. Judging the accuracy of the final content; D. Measuring whether the copied and pasted content is \textbf{useful and necessary} (tool-use proficiency).

The hard metrics, such as the success rate (CP S.R.), Rouge-L (R-L), and PPL in Table \ref{table:cp-bench-object}, focus on A, C, and B, respectively. While the GPT-based metrics such as clarity, accuracy, and consistency focus on B, C, and A, respectively. The hints are: 

\begin{itemize}
  \item \textbf{Clarity (B):} If the pasted contents are mislocated (B), the clarity will dramatically reduce due to increased spelling and grammatical errors.
  \item \textbf{Accuracy (C):} This refers to the correctness and precision of the information, as defined.
  \item \textbf{Consistency (A):} If the copy-paste tools are wrongly used, such as when incorrect contents are copied and pasted, the key information will be incorrect, leading to inconsistency in key information.
\end{itemize}

Note that these metrics are not orthogonal and can overlap with each other, but they each incline towards different aspects.

\end{document}